\documentclass[runningheads]{llncs}
\usepackage[toc,page]{appendix}
\usepackage{hyperref}
\usepackage{colortbl}
\usepackage[dvipsnames]{xcolor}
\usepackage{subcaption}
\usepackage{soul}
\usepackage{comment}
\usepackage{color}
\usepackage[toc,page]{appendix}
\usepackage{amsmath}
\usepackage{graphicx}
\usepackage{amsfonts}
\usepackage{bbm}
\usepackage{algorithm}
\usepackage{algpseudocode}
\usepackage{cite}
\newcommand{\floor}[1]{\left\lfloor #1 \right\rfloor}
\newcommand\numberthis{\addtocounter{equation}{1}\tag{\theequation}}

\newtheorem{assumption}{Assumption}

\usepackage[colorinlistoftodos,prependcaption]{todonotes}

\begin{document}
\title{Multi-Model Federated Learning \\ with Provable Guarantees}
%
%
\author{Neelkamal Bhuyan\inst{1} \and
Sharayu Moharir\inst{1} \and
Gauri Joshi\inst{2}}
\authorrunning{N. Bhuyan et al.}
%
\institute{Indian Institute of Technology Bombay, Mumbai, India \and Carnegie Mellon University, Pittsburgh, PA, USA}
\maketitle              
\begin{abstract}
Federated Learning (FL) is a variant of distributed learning where edge devices collaborate to learn a model without sharing their data with the central server or each other. We refer to the process of training multiple independent models simultaneously in a federated setting using a common pool of clients as multi-model FL. In this work, we propose two variants of the popular FedAvg algorithm for multi-model FL, with provable convergence guarantees. We further show that for the same amount of computation, multi-model FL can have better performance than training each model separately. We supplement our theoretical results with experiments in strongly convex, convex, and non-convex settings.


\keywords{Federated Learning  \and Distributed Learning \and Optimization}
\end{abstract}
\section{Introduction}
Federated Learning (FL) is a form of distributed learning where training is done by edge devices (also called clients) using their local data without sharing their data with the central coordinator (also known as server) or other devices. The only information shared between the clients and the server is the update to the global model after local client training. This helps in preserving the privacy of the local client dataset. Additionally, training on the edge devices does not require datasets to be communicated from the clients to the server, lowering communication costs.
The single model federated learning problem has been widely studied from various perspectives including optimizing the frequency of communication between the clients and the server \cite{MLSYS2019_c8ffe9a5}, designing client selection algorithms in the setting where only a subset of devices train the model in each iteration \cite{kairouz2021advances}, and measuring the effect of partial device participation on the convergence rate \cite{li2019convergence}.

Closest to this work, in \cite{bhuyan2022multi}, the possibility of using federated learning to train multiple models simultaneously using the same set of edge devices has been explored. We refer to this as multi-model FL. In the setting considered in \cite{bhuyan2022multi}, each device can only train one model at a time. The algorithmic challenge is to determine which model each client will train in any given round. Simulations illustrate that multiple models can indeed be trained simultaneously without a sharp drop in accuracy by splitting the clients into subsets in each round and use one subset to train each of the models. One key limitation of \cite{bhuyan2022multi} is the lack of analytical performance guarantees for the proposed algorithms. In this work, we address this limitation of \cite{bhuyan2022multi} by proposing algorithms for multi-model FL with provable performance guarantees. 

\subsection{Our Contributions} In this work, we focus on the task of training $M$ models simultaneously using a shared set of clients. We propose two variants of the Fed-Avg algorithm \cite{mcmahan2017communication} for the multi-model setting, with provable convergence guarantees. The first variant called Multi-FedAvg-Random (MFA-Rand) partitions clients into $M$ groups in each round and matches the $M$ groups to the $M$ models in a uniformly randomized fashion. Under the second variant called Multi-FedAvg-Round Robin (MFA-RR), time is divided into frames such that each frame consists of $M$ rounds. At the beginning of each frame, clients are partitioned into $M$ groups uniformly at random. Each group then trains the $M$ models in the $M$ rounds in the frame in a round-robin manner.

The error for a candidate multi-model federated learning algorithm $\mathcal{P}$, for each model is defined as the distance between each model's global weight at round $t$ and its optimizer. Our analytical results show that when the objective function is strongly convex and smooth, for MFA-Rand, an upper bound on the error decays as $\mathcal{O}(1/\sqrt{T})$, and for MFA-RR, the error decays as $\mathcal{O}(1/T)$. The latter result holds when local stochastic gradient descent at the clients transitions to full gradient descent over time. This allows MFA-RR to be considerably faster than FedCluster\cite{9377960}, an algorithm similar to MFA-RR.

Further, we study the performance gain in multi-model FL over single-model training under MFA-Rand. We show via theoretical analysis that training multiple models simultaneously can be more efficient than training them sequentially, i.e., training only one model at a time on the same set of clients. 

Via synthetic simulations we show that MFA-RR typically outperforms MFA-Rand. Intuitively this is because under MFA-RR, each client trains each model in every frame, while this is not necessary in the MFA-Rand. Our data-driven experiments prove that training multiple models simultaneously is advantageous over training them sequentially.


\subsection{Related Work} 
The federated learning framework and the FedAvg algorithm was first introduced in \cite{mcmahan2017communication}. Convergence of FedAvg has been studies extensively under convexity assumption \cite{koloskova2020unified,li2019convergence,woodworth2020minibatch} and non-convex setting \cite{koloskova2020unified,li2019communication}. When data is homogeneous across clients and all clients participate in each round, FedAvg is known as LocalSGD. LocalSGD is easier to analyse because of the aforementioned assumptions. LocalSGD has been proved to be convergent with strictly less communication in \cite{stich2018local}. LocalSGD has been proved to be convergent in non-convex setting \cite{coppola2015iterative, zhou2017convergence, wang2018cooperative}.

FedProx \cite{li2020federated} handles data heterogeneity among clients and convergences in the setting where the data across clients is non-iid. The key difference between FedAvg and FedProx is that the latter adds an addition proximal term to the loss function. However, the analysis of FedProx requires the proximal term to be present and, therefore, does not serve as proof for convergence of FedAvg.

Personalised FL is an area where multiple local models are trained instead of a global model. The purpose of having multiple models \cite{eichner2019semi} is to have the local models reflect the specific characteristics of their data. \cite{hanzely2020federated, deng2020adaptive, agarwal2020federated} involved using a combination of global and local models. Although there are multiple models, the underlying objective is same. This differentiates personalised FL from our setting, which involves multiple \emph{unrelated} models.

Clustered Federated Learning proposed in \cite{sattler2020clustered} involved performing FedAvg and  bi-partitioning client set in turn until convergence is reached. The idea here was to first perform vanilla FL. Then, recursively bi-partition the set of clients into two clusters and perform within the clusters until a more refined stopping criterion is met. The second step is a post-processing technique. \cite{castiglia2020multi} studies distributed SGD in heterogeneous networks, where distributed SGD is performed within sub-networks, in a decentralized manner, and model averaging happens periodically among neighbouring sub-networks.

FedCluster proposed in \cite{9377960} is very similar to second algorithm we propose in this work. In \cite{9377960}, convergence has been guaranteed in a non-convex setting. However, the advantages of a convex loss function or the effect of data sample size has not been explored in this work. Further, the clusters, here, are fixed throughout the learning process.

Training multiple models in a federated setting has been explored in \cite{9685230} also. However, the setting and the model training methodology are different from ours \cite{bhuyan2022multi}. In \cite{9685230}, client distribution among models has been approached as an optimization problem and a heuristic algorithm has been proposed. However, neither an optimal algorithm is provided nor any convergence guarantees.

\section{Setting}
We consider the setting where $M$ models are trained simultaneously in a federated fashion. The server has global version of each of the $M$ models and the clients have local datasets (possibly heterogeneous) for every model. The total number of  clients available for training is denoted by $N$. We consider full device participation with the set of clients divided equally among the $M$ models in each round. In every round, each client is assigned \emph{exactly one model} to train. Each client receives the current global weights of the model it was assigned and performs $E$ iterations of stochastic gradient descent locally at a fixed learning rate of $\eta_t$ (which can change across rounds indexed by $t$) using the local dataset. It then sends the local update to its model, back to the server. The server then takes an average of the received weight updates and uses it to get the global model for the next round.

Algorithm \ref{alg1} details the training process of multi-model FL with any client selection algorithm $\mathcal{P}$. Table \ref{tab0} lists out the variables used in the training process and their purpose.

\begin{table}
\caption{Common variables used in algorithms}
\begin{center}
\begin{tabular}{|c|c|}
\hline
\textbf{Variable Name} & \textbf{Description} \\
\hline
$n$ & round number\\
$M$ & Number of Models\\
$\mathcal{S}_T$ & Set of all clients \\
$N$ & Total number of clients = $|\mathcal{S}_T|$\\
$\mathcal{P}$ & Client selection algorithm\\
trainModel[$c$] & Model assigned to client $c$\\
globalModel[$m$] & Global weights of model $m$\\
localModel[$m,c$] & Weights of $m^{\text{th}}$  model of client $c$\\
\hline
\end{tabular}
\label{tab0}
\end{center}
\end{table}

\begin{algorithm}
\caption{Pseudo-code for $M$-model training at server}\label{alg1}
\hspace*{\algorithmicindent} \textbf{Input:} $\mathcal{S}_T$, algorithm $\mathcal{P}$\\
\hspace*{\algorithmicindent} \textbf{Initialize:} globalModel, localModel
\begin{algorithmic}[1]
\State globalModel$[m]$ $\gets$ \textbf{0} $\forall$ $m \in \{1,2,..,M\}$
\State Initialise parameters relevant to algorithm $\mathcal{P}$ in round 0
\Repeat
\State localModel$[m,c]$ $\gets$ \textbf{0} $\forall$ $c \in \mathcal{S}_T, m \in \{1,2,..,M\}$
\State Update parameters relevant to algorithm $\mathcal{P}$ 
\State trainModel $\gets$ call function for $\mathcal{P}$ \Comment{MFA\_Rand(t) or MFA\_RR(t)}
\For{$c \in \mathcal{S}_T$}
    \State $m$ $\gets$ trainModel$[c]$
    \State localModel$[m,c]$ $\gets$ globalModel$[m]$
    \State localUpdate[m,c] $\gets$ update  for  localModel$[m,c]$ after $E$ iterations of gradient descent
\EndFor
\For{$m \in \{1,2,..,M\}$}
    \State globalUpdate[m] $\gets$ weighted average of localUpdate[m,c]
    \State globalModel$[m]$ $\gets$ globalModel$[m]$ - globalUpdate[m]
\EndFor
\Until{$n$ = max iterations}
\end{algorithmic}
\end{algorithm}

The global loss function of the $m^{\text{th}}$ model is denoted by $F^{(m)}(\textbf{w})$, the local loss function of the $m^{\text{th}}$ model for the $k^{\text{th}}$ client is denoted by $F_k^{(m)}(\textbf{w})$.
For simplicity, we assume that, for every model, each client has same number of data samples, $\mathcal{N}$.
We therefore have the following global objective function for $m^{\text{th}}$ model,
\begin{equation}\label{eqn1}
    F^{(m)}(\textbf{w}) 
    =  \frac{1}{N}\sum_{k=1}^N F_{k}^{(m)}(\textbf{w}).
\end{equation}
Additionally, $\Gamma^{(m)} = \min (F^{(m)}) - \frac{1}{N}\sum_{k=1}^N \min(F_k^{(m)}) \ge 0$.
We make some standard assumptions on the local loss function, also used in \cite{cho2020client,li2019convergence}. These are,

\begin{assumption}\label{assump1}
All $F_k^{(m)}$ are $\mu$-strongly convex.
\end{assumption}
\begin{assumption}\label{assump2}
All $F_k^{(m)}$ are $L$-smooth.
\end{assumption}

\begin{assumption}\label{assump3}
Stochastic gradients are bounded: $\mathbb{E}||\nabla F_k^{(m)}(\textbf{w}, \xi)||^2 \le (G^{(m)})^2$.
\end{assumption}

\begin{assumption}\label{assump4}
For each client k, $F_k^{(m)}(\textbf{\textbf{w}}) = \frac{1}{\mathcal{N}} \displaystyle \sum_{y=1}^\mathcal{N} f_{k,y}^{(m)}(\textbf{w})$, where $f_{k,y}^{(m)}$ is the loss function of the $y^{th}$ data point of the $k^{th}$ client's $m^{th}$ model. Then for each $f_{k,y}^{(m)}$,
\begin{equation*}
    ||\nabla f_{k,y}^{(m)}(\textbf{w})||^2 \le \beta_1^{(m)} + \beta_2^{(m)}||\nabla F_k^{(m)}(\textbf{w})||^2,
\end{equation*}
for some constants $\beta_1^{(m)} \ge 0$ and $\beta_2^{(m)} \ge 1$.
\end{assumption}
The last assumption is standard in the stochastic optimization literature \cite{bertsekas1995neuro, friedlander2012hybrid}.

\subsection{Performance Metric}
The error for a candidate multi-model federated learning algorithm $\mathcal{P}$'s is the distance between the $m^{\text{th}}$ model's global weight at round $t$, denoted by $\overline{\textbf{w}}^{(m)}_{\mathcal{P},t}$ and the minimizer of $F^{(m)}$, denoted by $\textbf{w}^{(m)}_*$. Formally,
\begin{equation}\label{eqn2}
    \Delta^{(m)}_{\mathcal{P}}(t) = \mathbb{E}||\overline{\textbf{w}}^{(m)}_{\mathcal{P},t} - \textbf{w}^{(m)}_*||.
\end{equation}

\section{Algorithms}
We consider two variants of the Multi-FedAvg algorithm proposed in \cite{bhuyan2022multi}. For convenience we assume that $N$ is an integral multiple of $M$.

\subsection{Multi-FedAvg-Random (MFA-Rand)}
The first variant partitions the set of clients $\mathcal{S}_T$ into $M$ equal sized subsets $\{ \mathcal{S}_1^t, \mathcal{S}_2^t,...,\mathcal{S}_M^t \}$ in every round $t$. The subsets are created uniformly at random independent of all past and future choices. The $M$ subsets are then matched to the $M$ models with the matching chosen uniformly at random.

Algorithm \ref{alg2} details the sub-process of MFA-Rand invoked when client-model assignment step (step 6) runs in Algorithm \ref{alg1}. An example involving 3 models over 6 rounds has been worked out in Table \ref{tab:tab3}. 


\begin{table}
    \centering
    \caption{MFA-Rand over 6 rounds for 3 models. }
    \begin{tabular}{|c|c|c|c|}
        \hline
        \textbf{Round} & \textbf{Model 1} & \textbf{Model 2} & \textbf{Model 3} \\
        \hline
        1 & $\mathcal{S}_1^1$ & $\mathcal{S}_2^1$ & $\mathcal{S}_3^1$ \\
        \hline
        2 & $\mathcal{S}_1^2$ & $\mathcal{S}_2^2$ & $\mathcal{S}_3^2$ \\
        \hline
        3 & $\mathcal{S}_1^3$ & $\mathcal{S}_2^3$ & $\mathcal{S}_3^3$ \\
        \hline
        4 & $\mathcal{S}_1^4$ & $\mathcal{S}_2^4$ & $\mathcal{S}_3^4$ \\
        \hline
        5 & $\mathcal{S}_1^5$ & $\mathcal{S}_2^5$ & $\mathcal{S}_3^5$ \\
        \hline
        6 & $\mathcal{S}_1^6$ & $\mathcal{S}_2^6$ & $\mathcal{S}_3^6$ \\
        \hline
    \end{tabular} 
    \label{tab:tab3}
\end{table}

\begin{algorithm}
\caption{Pseudo-code for MFA-Rand}\label{alg2}


\begin{algorithmic}[1]
\Procedure{MFA\_Rand}{$t$}
\State $\{\mathcal{S}_1^t, \mathcal{S}_2^t,..., \mathcal{S}_M^t\} \gets $Partition $\mathcal{S}_T$ randomly into $M$ disjoint subsets 
\For{$\mathcal{S} \in \{\mathcal{S}_1^t, \mathcal{S}_2^t,..., \mathcal{S}_M^t\}$}
\If{$\mathcal{S} = \mathcal{S}_j^t$}
\State trainModel[$c$] $\gets$ $j$ $\forall$ $c \in \mathcal{S}$
\EndIf
\EndFor
\State \textbf{return:} trainModel
\EndProcedure
\end{algorithmic}
\end{algorithm}

\subsection{Multi-FedAvg-Round Robin (MFA-RR)} The second variant partitions the set of clients into $M$ equal sized subsets once every $M$ rounds. The subsets are created uniformly at random independent of all past and future choices. We refer to the block of $M$ rounds during which the partitions remains unchanged as a frame. Within each frame, each of the $M$ subsets is mapped to each model exactly once in a round-robin manner. Specifically, let the subsets created at the beginning of frame $l$ be denoted by $\mathcal{S}_j^{l}$ for $1 \leq j \leq M$. Then, in the $u^{\text{th}}$ round in frame $l$ for $1 \leq u \leq M$, $\mathcal{S}_j^{l}$ is matched to model $((j+u-2) \mod M) + 1$.

Algorithm \ref{alg3} details the sub-process of MFA-RR invoked when client-model assignment step (step 6) runs in Algorithm \ref{alg1}. An example involving 3 models over 6 rounds has been worked out in Table \ref{tab:tab4}. 

\begin{table}
    \centering
    \caption{MFA-RR over 6 rounds (2 frames) for 3 models.}
    \begin{tabular}{|c|c|c|c|}
        \hline
        \textbf{Round} & \textbf{Model 1} & \textbf{Model 2} & \textbf{Model 3} \\
        \hline
        1 & \cellcolor{Goldenrod} $\mathcal{S}_1^1$ & \cellcolor{LimeGreen}$\mathcal{S}_2^1$ & \cellcolor{SkyBlue}$\mathcal{S}_3^1$ \\
        \hline
        2 & \cellcolor{SkyBlue}$\mathcal{S}_3^1$ & \cellcolor{Goldenrod}$\mathcal{S}_1^1$ & \cellcolor{LimeGreen}$\mathcal{S}_2^1$ \\
        \hline
        3 & \cellcolor{LimeGreen}$\mathcal{S}_2^1$ & \cellcolor{SkyBlue}$\mathcal{S}_3^1$ & \cellcolor{Goldenrod}$\mathcal{S}_1^1$ \\
        \hline
        4 & \cellcolor{Magenta}$\mathcal{S}_1^2$ & \cellcolor{Gray}$\mathcal{S}_2^2$ & \cellcolor{Orange}$\mathcal{S}_3^2$ \\
        \hline
        5 & \cellcolor{Orange}$\mathcal{S}_3^2$ & \cellcolor{Magenta}$\mathcal{S}_1^2$ & \cellcolor{Gray}$\mathcal{S}_2^2$ \\
        \hline
        6 & \cellcolor{Gray}$\mathcal{S}_2^2$ & \cellcolor{Orange}$\mathcal{S}_3^2$ & \cellcolor{Magenta}$\mathcal{S}_1^2$ \\
        \hline
    \end{tabular} 
    \label{tab:tab4}
\end{table}

\begin{algorithm}
\caption{Pseudo-code for MFA-RR}\label{alg3}

\begin{algorithmic}[1]
\Procedure{MFA\_RR}{$t$}
\If{$t \mod M = 1$}
\State $l = \frac{t-1}{M}+1$
\State $\{\mathcal{S}_1^l, \mathcal{S}_2^l,..., \mathcal{S}_M^l\} \gets $ Partition $\mathcal{S}_T$ randomly into $M$ disjoint subsets
\EndIf
\For{$\mathcal{S} \in \{\mathcal{S}_1^l, \mathcal{S}_2^l,..., \mathcal{S}_M^l\}$}

\If{$\mathcal{S} = \mathcal{S}_j^l$}
\State trainModel[$c$] $\gets$ $((j+t-2) \mod M) + 1$ $\forall$ $c \in \mathcal{S}$
\EndIf
\EndFor
\State \textbf{return:} trainModel
\EndProcedure
\end{algorithmic}
\end{algorithm}

\begin{remark}
An important difference between the two algorithms is that
under MFA-RR, each client-model pair is used exactly once in each frame consisting of $M$ rounds, whereas under MFA-Rand, a specific client-model pair is not matched in $M$ consecutive time-slots with probability $\left(1-\frac{1}{M}\right)^M$ .
\end{remark}

\section{Convergence of MFA-Rand and MFA-RR}
In this section, we present our analytical results for MFA-Rand and MFA-RR.
\begin{theorem}
For MFA-Rand, under Assumptions \ref{assump1}, \ref{assump2}, \ref{assump3}, \ref{assump4} and with $\eta_t = \frac{\beta}{t + \gamma}$, where $\beta > \frac{1}{\mu}$ and $\gamma > 4L\beta - 1$, we have
\begin{equation*}\label{thm1}
    \Delta_{\text{MFA-Rand}}^{(m)}(t) \le \frac{\sqrt{\nu}}{\sqrt{t+\gamma}} \text{ } M\ge 1 \text{ and } E\ge 1,
\end{equation*}
where $\nu = \max\left\{\frac{\beta^2 (B+C)}{\beta\mu-1},\mathbb{E}||\overline{\textbf{w}}^{(m)}_{\text{MFA-Rand},1} - \textbf{w}^{(m)}_*||^2(1+\gamma)\right\}, B = 6L\Gamma^{(m)} + (1/N^2)\sum_{k=1}^N (\sigma_k^{(m)})^2 + 8(E-1)^2 (G^{(m)})^2, C = \frac{M-1}{N-1}E^2 (G^{(m)})^2$ and $(\sigma_k^{(m)})^2 = 4(\beta_1 + \beta_2 (G^{(m)})^2)$.
\end{theorem}
The result follows from the convergence of FedAvg with partial device participation in \cite{li2019convergence}. Note that $B$ and $C$ influence the lower bound on number of iterations, $T_{\text{MFA-Rand}}(\epsilon)$, required to reach a certain accuracy. Increasing the number of models, $M$, increases $C$ which increases $T_{\text{MFA-Rand}}(\epsilon)$.

\begin{theorem}\label{thm2}
Consider MFA-RR, under Assumptions \ref{assump1}, \ref{assump2}, \ref{assump3}, \ref{assump4} and $\eta_t = \frac{\beta}{1 + \floor{\frac{t}{M}} + \gamma}$, where $\beta > \frac{1}{\mu}$ and $\gamma > \beta L - 1$. Further, $\mathcal{N}_s(t)$ is the sample size for SGD iterations at clients in round $t$. If 
\begin{equation*}
    \frac{\mathcal{N} - \mathcal{N}_s(t)}{\mathcal{N}} \le \eta_t \left(\frac{V}{2E  \sqrt{\beta_1 + \beta_2 G^2}} \right) 
\end{equation*}
for some $V \ge 0$, $\beta_1 = \displaystyle \max_{m} \beta_1^{(m)}$ and $\beta_2 = \displaystyle \max_{m} \beta_2^{(m)}$, then,
\begin{equation*}
    \Delta_{\text{MFA-RR}}^{(m)}(t) \le \frac{\phi}{\frac{t}{M} + \gamma} \text{ } \forall \text{ } M\ge 1 \text{ and } E\ge 1,
\end{equation*}
where $\phi = \frac{\beta E G^{(m)}(M-1)}{M} + \max \left\{\frac{\beta^2(Y+Z+V)}{\beta \mu -1}, (1+\gamma)\Delta^{(m)}_{\text{MFA-RR}}(1) \right\},$\\ $Y = \frac{LG^{(m)} E^2 (M-1)}{2M}$ and $Z = LG^{(m)} E(E-1)$.
\end{theorem}

Here, we require that the SGD iterations at clients have data sample converging sub-linearly to the full dataset. For convergence, the only requirement is $\frac{\mathcal{N} - \mathcal{N}_s(t)}{\mathcal{N}}$ to be proportional to the above defined $\eta_t$.

We observe that $Y$, $Z$ and $V$ influence the lower bound on number of iterations, $T_{\text{MFA-RR}}(\epsilon)$, required to reach a certain accuracy. Increasing the number of models, $M$, increases $Y$ which increases $T_{\text{MFA-RR}}(\epsilon)$.

A special case of Theorem \ref{thm2} is when we employ full gradient descent instead of SGD at clients. Naturally, in this case, $V = 0$. 
\begin{corollary}\label{corr1}
For MFA-RR with full gradient descent at the clients, under Assumptions \ref{assump1}, \ref{assump2}, \ref{assump3}, \ref{assump4} and $\eta_t = \frac{\beta}{1+\floor{\frac{t}{M}} + \gamma}$, 
\begin{equation*}
    \Delta_{\text{MFA-RR}}^{(m)}(t) \le \frac{\phi'}{\frac{t}{M} + \gamma} \text{ } \forall \text{ } M\ge 1 \text{ and } E\ge 1,
\end{equation*}
where $\phi' = \frac{\beta E G^{(m)}(M-1)}{M} + \max \left\{\frac{\beta^2(Y+Z)}{\beta \mu -1}, (1+\gamma)\Delta^{(m)}_{\text{MFA-RR}}(1) \right\}$,\\ $Y = \frac{LG^{(m)} E^2 (M-1)}{2M}$ and $Z = LG^{(m)} E(E-1)$.
\end{corollary}

\begin{remark}
MFA-RR, when viewed from perspective of one of the $M$ models, is very similar to FedCluster. However, there are some key differences between the analytical results. 

First, FedCluster assumes that SGD at client has fixed bounded variance for any sample size (along with Assumption \ref{assump3}). This is different from Assumption \ref{assump4} of ours. When Assumption \ref{assump4} is coupled with Assumption \ref{assump3}, we get sample size dependent bound on the variance. A smaller variance is naturally expected for a larger data sample. Therefore, our assumption is less restrictive.

Second, the effect of increasing sample size (or full sample) has not been studied in \cite{9377960}. We also see the effect of strong convexity on the speed of convergence. The convergence result from \cite{9377960} is as follows,
\begin{equation*}
    \frac{1}{T} \sum_{j=0}^{T-1} \mathbb{E}||\nabla F(\overline{\textbf{w}}_{jM})||^2 \le \mathcal{O}\left(\frac{1}{\sqrt{T}}\right).
\end{equation*}
If we apply the strong convexity assumption to this result and use that $\mu ||x - x_*|| \le ||\nabla F(x)||$, we get
\begin{equation*}
    \frac{\mu^2}{T} \sum_{j=0}^{T-1} \mathbb{E}||\overline{\textbf{w}}_{jM} - \textbf{w}_*||^2 \le \mathcal{O}\left(\frac{1}{\sqrt{T}}\right).
\end{equation*}
Applying Cauchy-Schwartz inequality on LHS we get,
\begin{equation*}
    \left(\frac{1}{T} \sum_{j=0}^{T-1} \mathbb{E}||\overline{\textbf{w}}_{jM} - \textbf{w}_*||\right)^2 \le \frac{1}{T} \sum_{j=0}^{T-1} \mathbb{E}||\overline{\textbf{w}}_{jM} - \textbf{w}_*||^2.
\end{equation*}
Finally, we have
\begin{equation*}
    \frac{1}{T} \sum_{j=0}^{T-1} \mathbb{E}||\overline{\textbf{w}}_{jM} - \textbf{w}_*|| \le \mathcal{O}\left(\frac{1}{T^{1/4}}\right),
\end{equation*}
for any sampling strategy. With an increasing sample size (or full sample size), we can obtain $\mathcal{O}(1/T)$ convergence. This is a significant improvement over the convergence result of FedCluster.
\end{remark}

\section{Advantage of Multi-Model FL over single model FL}
We quantify the advantage of Multi-Model FL over single model FL by defining the gain of a candidate  multi-model Federated Learning algorithm $\mathcal{P}$ over FedAvg \cite{mcmahan2017communication}, which trains only one model at a time.\\
Let $T_1(\epsilon)$ be the number of rounds needed by one of the $M$ models using FedAvg to reach an accuracy level (distance of model's current weight from its optimizer) of $\epsilon$. We assume that all $M$ models are of similar complexity. This means we expect that each model reaches the required accuracy in roughly the same number of rounds. Therefore, cumulative number of rounds needed to ensure all $M$ models reach an accuracy level of $\epsilon$ using FedAvg is $MT_1(\epsilon)$. Further, let the number of rounds needed to reach an accuracy level of $\epsilon$ for all $M$ models under $\mathcal{P}$ be denoted by $T_{\mathcal{P}}(M,\epsilon)$. We define the gain of algorithm $\mathcal{P}$ for a given $\epsilon$ as
\begin{equation}\label{eqn10}
    g_{\mathcal{P}}(M, \epsilon) = \frac{MT_1(\epsilon)}{T_{\mathcal{P}}(M,\epsilon)}.
\end{equation}
Note that FedAvg and $\mathcal{P}$ use the same number of clients in each round, thus the comparison is fair. Further, we use the bounds in Theorem \ref{thm1} and Theorem \ref{thm2} as proxies for calculating $T_{\mathcal{P}}(M,\epsilon)$ for MFA-Rand and MFA-RR respectively.

\begin{theorem}\label{thm3}
When $\epsilon < \displaystyle \min_{m} \Delta_{\text{MFA-Rand}}^{(m)}(1)$ and $M \le \frac{N}{2}$, the following holds for the gain of $M$-model MFA-Rand over running FedAvg $M$ times
\begin{align*}
    g_{\text{MFA-Rand}}(M, \epsilon) &> 1 \text{ } \forall \text{ } M>1\\
    \frac{d}{dM} g_{\text{MFA-Rand}}(M, \epsilon) &> 0 \text{ } \forall \text{ } M \ge 1
\end{align*}
for all $E \ge 2$ and for $E=1$ when $N > 1 + \displaystyle \max_{m} \left \{ \frac{6L\Gamma^{(m)}}{(G^{(m)})^2}\right \}$.
\end{theorem}
We get that gain increases with $M$ upto $M = N/2$, after which we have the $M=N$ case. At $M = N$, each model is trained by only one client, which is too low, especially when $N$ is large.

For the case of $E=1$, Theorem \ref{thm3} puts a lower bound on $N$. However, the $E=1$ case is rarely used in practice. One of the main advantages of FL is the lower communication cost due to local training. This benefit is not utilised when $E=1$.

\section{Simulations in strongly convex setting}
\subsection{Simulation Framework}
We take inspiration from the framework presented in \cite{li2019convergence} where a quadratic loss function, which is strongly convex, is minimized in a federated setting. We employ MFA-Rand and MFA-RR algorithms and compare their performance in this strongly convex setting. In addition to that, we also measure the gain of MFA-Rand and MFA-RR over sequentially running FedAvg in this strongly convex setting.\\
The global loss function is
\begin{equation}\label{eqn3}
    F(w) = \frac{1}{2N}\left(\textbf{w}^T\textbf{A}\textbf{w} - 2\textbf{b}^T\textbf{w}\right) + \frac{\mu}{2}||\textbf{w}||^2,
\end{equation}
where $N>1$, $\textbf{A} \in \mathcal{\textbf{R}}^{(Np+1)\times(Np+1)}$, $\textbf{w} \in \mathcal{\textbf{R}}^{(Np+1)}$ and $\mu > 0$.
We have,
\begin{equation}\label{eqn4}
    \textbf{A} = \sum_{k=1}^N \textbf{A}_k
\end{equation}
\begin{equation}\label{eqn5}
    \textbf{b} = \sum_{k=1}^N \textbf{b}_k.
\end{equation}
Here the $k^{th}$ pair $(\textbf{A}_k, \textbf{b}_k)$ represents the $k^{th}$ client.
Following is the definition of $\textbf{A}_k$
\begin{equation}\label{eqn6}
    \textbf{A}_k = \begin{cases} 
    \textbf{B}_k + \textbf{E}_{1,1} & k=1\\
    \textbf{B}_k & 1<k<N\\
    \textbf{B}_k + \textbf{E}_{Np+1,Np+1} & k=N,\end{cases}
\end{equation}
where $\textbf{E}_{i,j}$ is a matrix where $(i,j)$ element is 1 and rest all are 0. $\textbf{B}_k$ is defined as follows

\[
\left (
\begin{array}{ccc}
\textbf{0}_{(k-1)p\times (k-1)p} & &\\
& \left(\begin{array}{ccccc}
    1 & -1 & & & \\
    -1 & 2 & -1 & & \\
     & -1 & \ddots & \ddots & \\
     & & \ddots & 2 & -1\\
     & & & -1 & 1
\end{array}\right) & \\
& & \textbf{0}_{(N-k)p\times (N-k)p}
\end{array}
\right )
\]
and $\textbf{b}_k$ is defined as follows
\begin{equation}\label{eqn7}
    \textbf{b}_k = \begin{cases} 
    \textbf{e}_k & k=1\\
    \textbf{0 }& 1<k \le N. \end{cases}
\end{equation}
We finally define local loss function of the $k^{\text{th}}$ client as
\begin{equation}\label{eqn8}
    F_k(\textbf{w}) = \frac{1}{2}\left(\textbf{w}^T\textbf{A}_k\textbf{w} - 2\textbf{b}_k^T\textbf{w}\right) + \frac{\mu}{2}||\textbf{w}||^2,
\end{equation}
which satisfies our problem statement $F = \frac{1}{N}\sum_{k=1}^N F_k$.

\begin{remark}
We simulate the minimization of a single global loss function while talking about multi-model learning. The reason behind this is that we are doing these simulations from the perspective of one of the $M$ models. Therefore, $M$-model MFA-Rand boils down to sampling $N/M$ clients independently every round while $M$-model MFA-RR remains the same (going over subsets $\{\mathcal{S}_1^l, \mathcal{S}_2^l,.., \mathcal{S}_M^l\}$ in frame $l$). Furthermore, the gain simulations here assume that all $M$ models are of the same complexity.
\end{remark}

\subsection{Comparison of MFA-Rand and MFA-RR}
We consider the scenario where $N=24$, $p=4$ and $\mu = 2 \times 10^{-4}$. We take $E=5$, meaning 5 local SGD iterations at clients. We track the log distance of the current global loss from the global loss minimum, that is
\begin{equation}\label{eqn9}
    gap(t) = \log_{10}(F(w) - F(w_*))
\end{equation}
for 1000 rounds. We consider both constant $(\eta_t = 0.1)$ and decaying learning rate $\left(\eta_t = \frac{30}{100 + t}\right)$for $M=2$ and $M=12$.

\begin{figure}
     \centering
     \begin{subfigure}[b]{0.45\textwidth}
         \centering
         \includegraphics[width=\textwidth]{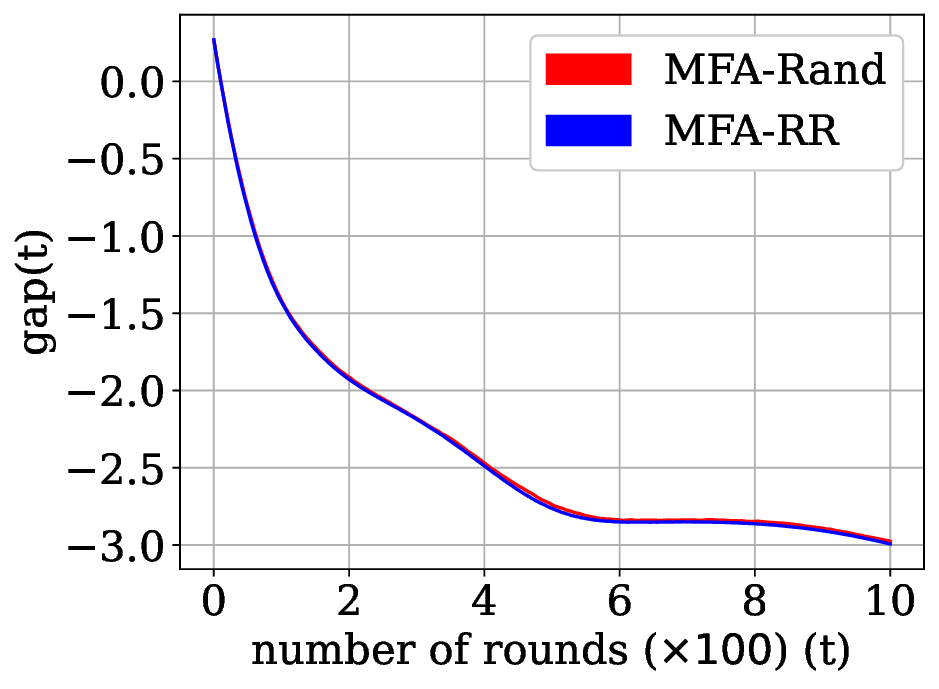}
         \caption{Averaged over 20 sample runs}
         \label{fig:1a}
     \end{subfigure}
     \hfill
     \begin{subfigure}[b]{0.45\textwidth}
         \centering
         \includegraphics[width=\textwidth]{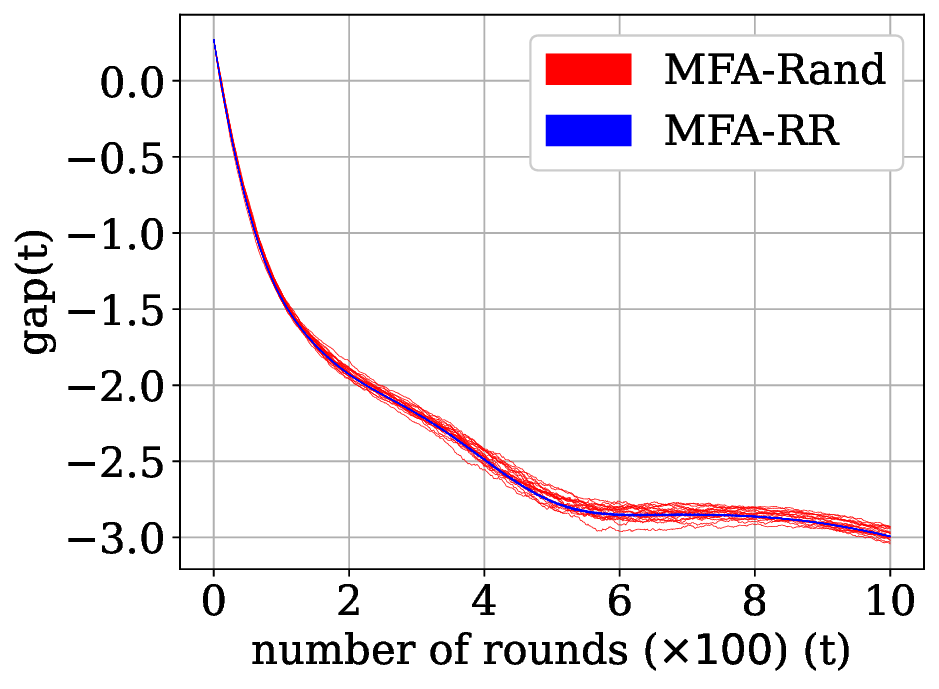}
         \caption{20 sample runs each}
         \label{fig:1b}
     \end{subfigure}
        \caption{$M=2$, constant learning rate}
        \label{fig:fig1}
\end{figure}
\begin{figure}
     \centering
     \begin{subfigure}[b]{0.45\textwidth}
         \centering
         \includegraphics[width=\textwidth]{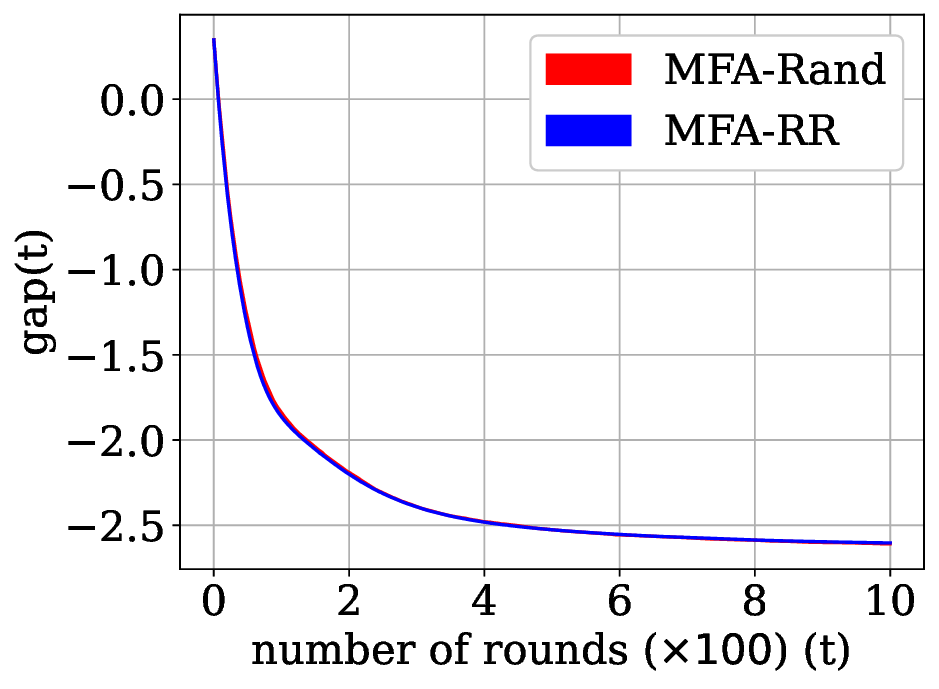}
         \caption{Averaged over 20 sample runs}
         \label{fig:2a}
     \end{subfigure}
     \hfill
     \begin{subfigure}[b]{0.45\textwidth}
         \centering
         \includegraphics[width=\textwidth]{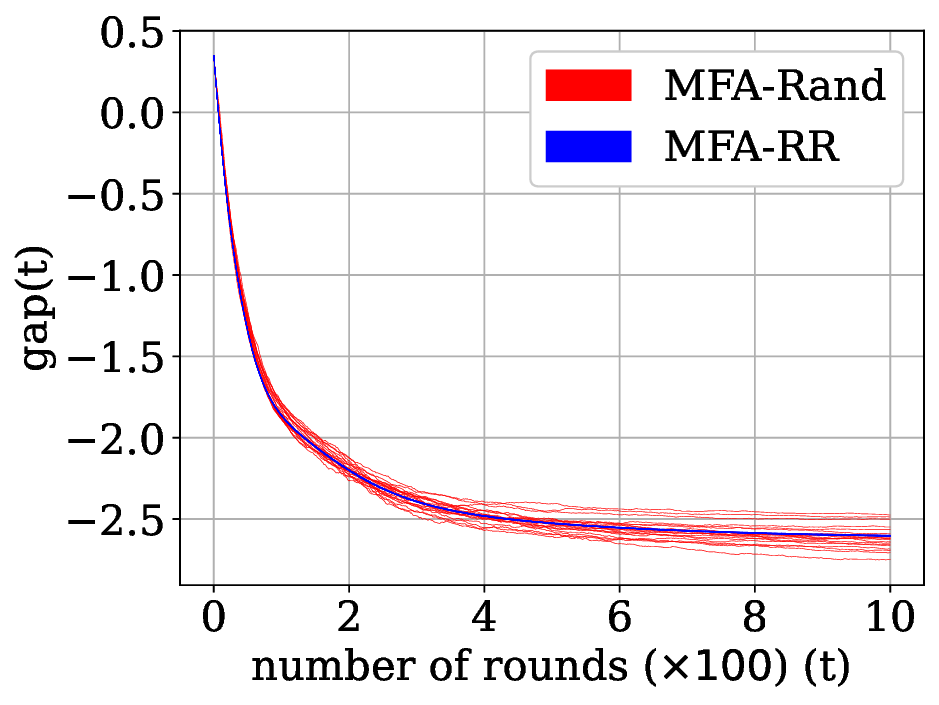}
         \caption{20 sample runs each}
         \label{fig:2b}
     \end{subfigure}
        \caption{$M=2$, decaying learning rate}
        \label{fig:fig2}
\end{figure}
As one can observe in Fig. \ref{fig:1a} and Fig. \ref{fig:2a}, we have similar mean performance for MFA-Rand and MFA-RR. However, Fig. \ref{fig:1b} and Fig. \ref{fig:2b} reveal that the randomness involved in MFA-Rand is considerably higher than that in MFA-RR, showing the latter to be more reliable.

\begin{figure}
     \centering
     \begin{subfigure}[b]{0.45\textwidth}
         \centering
         \includegraphics[width=\textwidth]{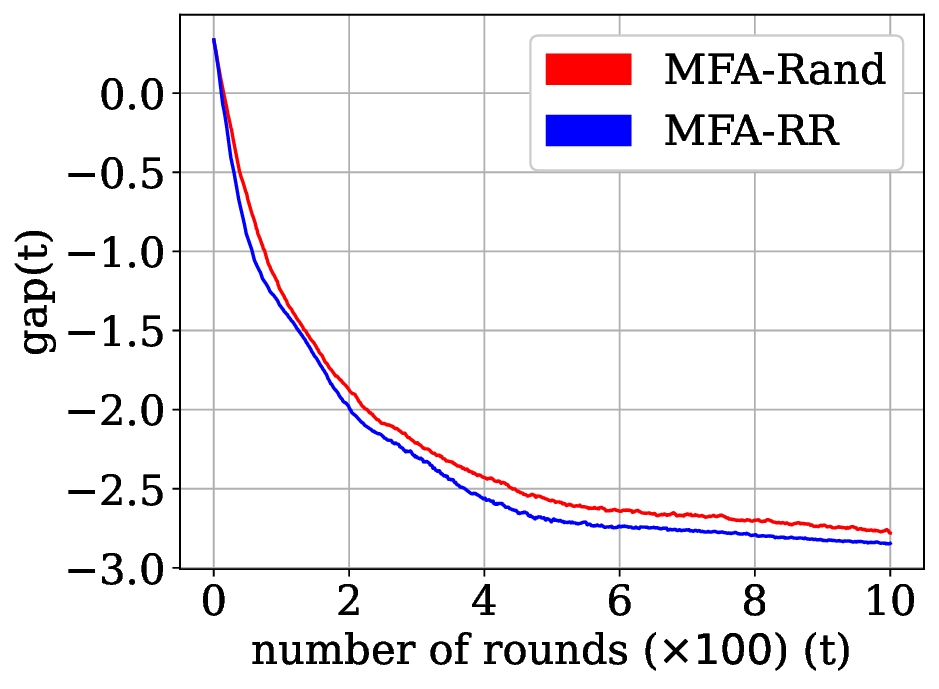}
         \caption{Averaged over 20 sample runs}
         \label{fig:3a}
     \end{subfigure}
     \hfill
     \begin{subfigure}[b]{0.45\textwidth}
         \centering
         \includegraphics[width=\textwidth]{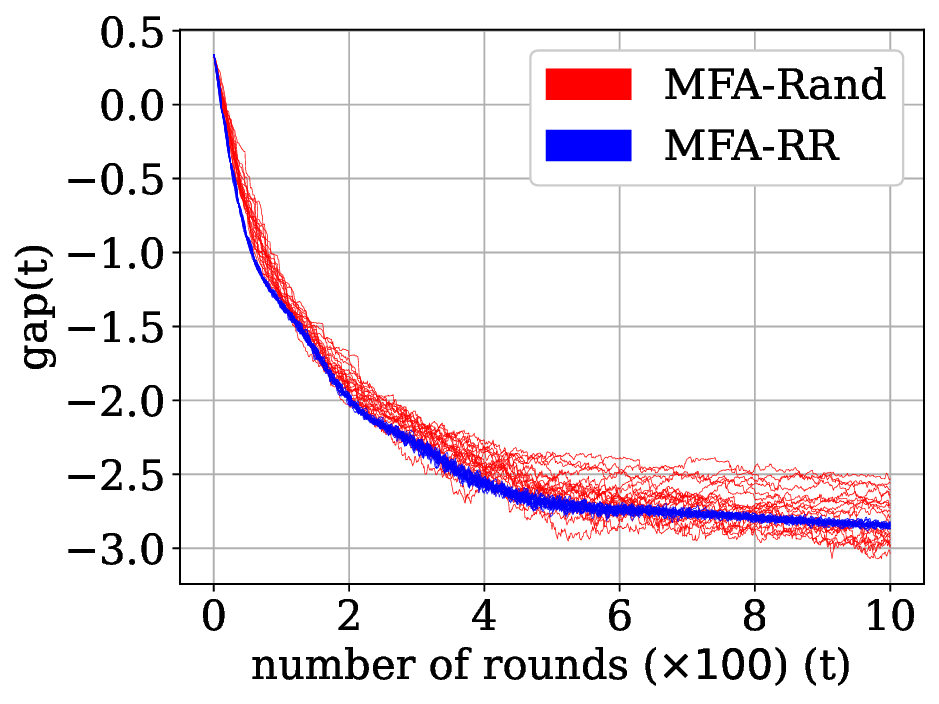}
         \caption{20 sample runs each}
         \label{fig:3b}
     \end{subfigure}
        \caption{$M=12$, constant learning rate}
        \label{fig:fig3}
\end{figure}
\begin{figure}
     \centering
     \begin{subfigure}[b]{0.45\textwidth}
         \centering
         \includegraphics[width=\textwidth]{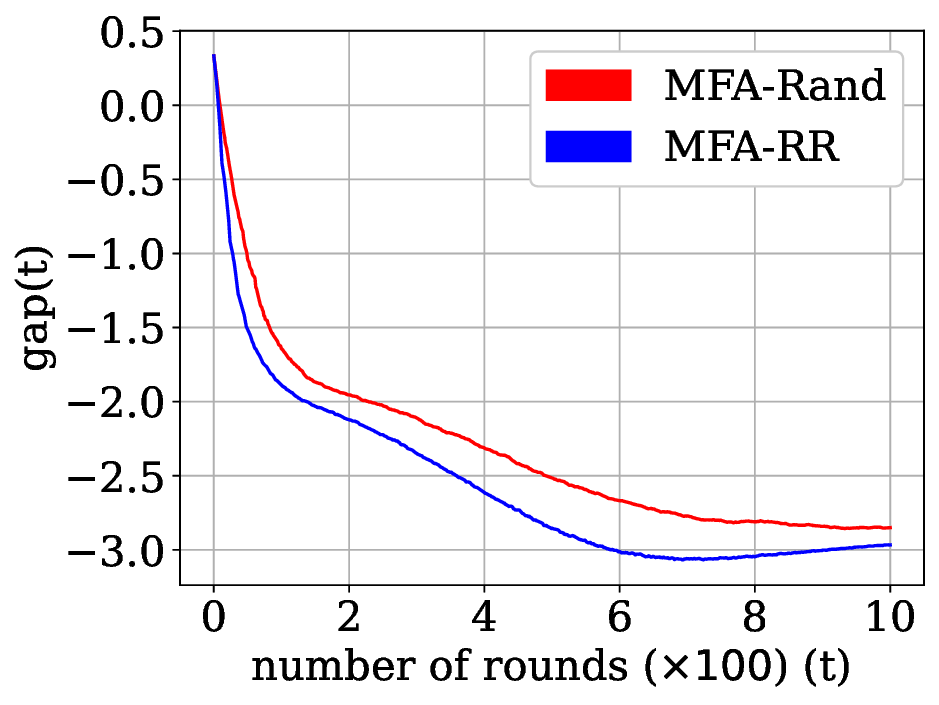}
         \caption{Averaged over 20 sample runs}
         \label{fig:4a}
     \end{subfigure}
     \hfill
     \begin{subfigure}[b]{0.45\textwidth}
         \centering
         \includegraphics[width=\textwidth]{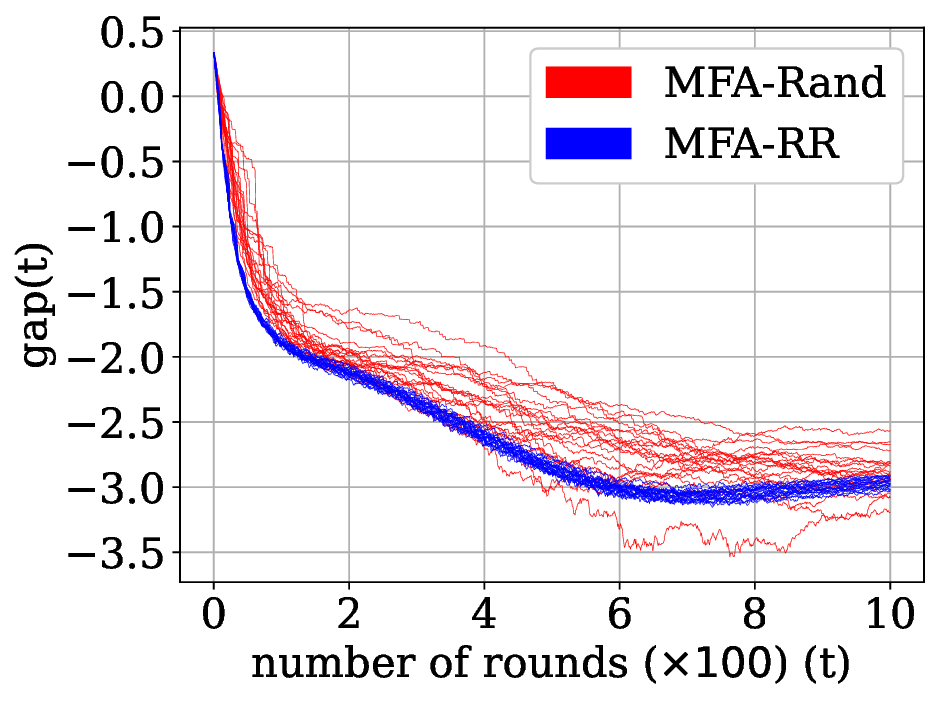}
         \caption{20 sample runs each}
         \label{fig:4b}
     \end{subfigure}
        \caption{$M=12$, decaying learning rate}
        \label{fig:fig4}
\end{figure}
It is evident from Fig. \ref{fig:3a} and Fig. \ref{fig:4a} that MFA-RR, on an average, performs better than MFA-Rand when $M$ is high. Again, Fig.\ref{fig:3b} and Fig. \ref{fig:4b} show that MFA-Rand has considerably higher variance than MFA-RR.

\begin{remark}
In this set of simulations, each client performs full gradient descent. While the analytical upper bounds on errors suggest an order-wise difference in the performance of MFA-RR and MFA-Rand, we do not observe that significant a difference between the performance of the two algorithms. This is likely because our analysis of MFA-RR exploits the fact that each client performs full gradient descent, while the analysis of MFA-Rand adapted from \cite{li2019convergence} does not.
\end{remark}

\subsection{Gain vs $M$}
\begin{figure}
    \centering
     \begin{subfigure}[b]{0.45\textwidth}
         \centering
         \includegraphics[width=\textwidth]{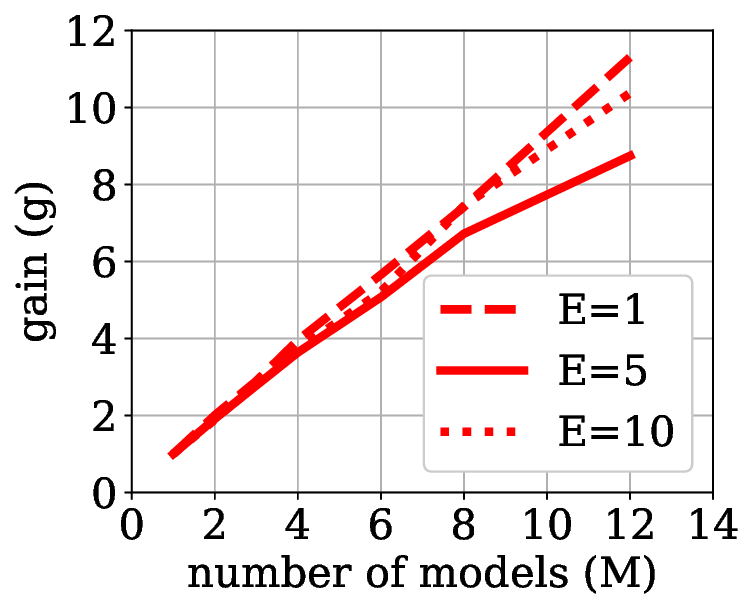}
         \caption{MFA-Rand}
         \label{fig:8a}
     \end{subfigure}
     \hfill
     \begin{subfigure}[b]{0.45\textwidth}
         \centering
         \includegraphics[width=\textwidth]{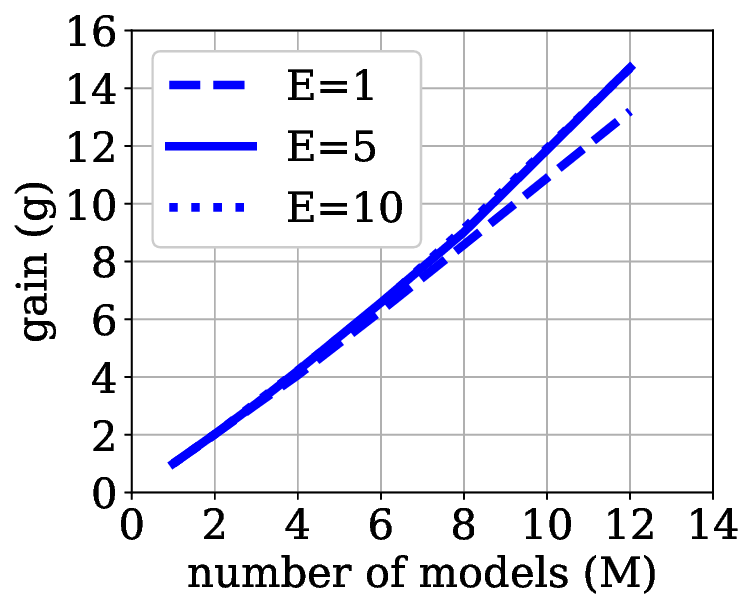}
         \caption{MFA-RR}
         \label{fig:8b}
     \end{subfigure}
        \caption{Gain vs $M$ in strongly convex setting}
        \label{fig:fig8}
\end{figure}

We test with $N=24$ clients for $M \leq 12$ for $E=1,5$ and $10$. Gain vs $M$ plots in Fig. \ref{fig:fig8} show that gain increases with $M$ for both MFA-Rand and MFA-RR.

\section{Data Driven Experiments on Gain}
We use Synthetic(1,1) \cite{li2020federated,li2019fair} and CelebA \cite{caldas2018leaf}\cite{liu2015deep} datasets for these experiments. The learning task in Synthetic(1,1) is multi-class logistic regression classification of feature vectors. Synthetic(1,1)-A involves 60 dimensional feature vectors classified into 5 classes while Synthetic(1,1)-B involves 30 dimensional feature vectors classified into 10 classes. CelebA dataset involves binary classification of face images based on a certain facial attribute, (for example, blond hair, smiling, etc) using convolutional neural networks (CNNs). The dataset has many options for the facial attribute.

The multi-model FL framework for training multiple unrelated models simultaneously was first introduced in our previous work \cite{bhuyan2022multi}. We use the same framework for these experiments. We first find the gain vs $M$ trend for Synthetic(1,1)-A, Synthetic(1,1)-B and CelebA. Then, we simulate a real-world scenarios where each of the $M$ models is a different learning task.

\subsection{Gain vs $M$} Here, instead of giving $M$ different tasks as the $M$ models, we have all $M$ models as the same underlying task. The framework, however, treats the $M$ models as independent of each other. This ensures that the $M$ models are of equal complexity.

We plot gain vs $M$ for two kinds of scenarios. First, when all clients are being used in a round. Theorem \ref{thm3} assumes this scenario. We call it full device participation as all clients are being used. Second, when only a sample, of the set of entire clients, is selected to be used in the round (and then distributed among the models). We call this partial device participation as a client has a non-zero probability of being idle during a round.

\subsubsection{Full device participation:}
For Synthetic(1,1)-A, we have $N=100$ clients and $T_1=70$. During single model FedAvg, we get 51\% training accuracy and 51\% test accuracy at the end of 70 rounds.

Synthetic(1,1)-B, we have $N=100$ clients and $T_1=70$. During single model FedAvg, we get 42.7\% training accuracy and 42.8\% test accuracy at the end of 70 rounds.

For CelebA, we have 96 clients and $T_1 = 75$. We get 79\% training accuracy and 68\% test accuracy at the end of 75 rounds of single model FedAvg.

\begin{figure}
     \centering
     \begin{subfigure}[b]{0.3\textwidth}
         \centering
         \includegraphics[width=\textwidth]{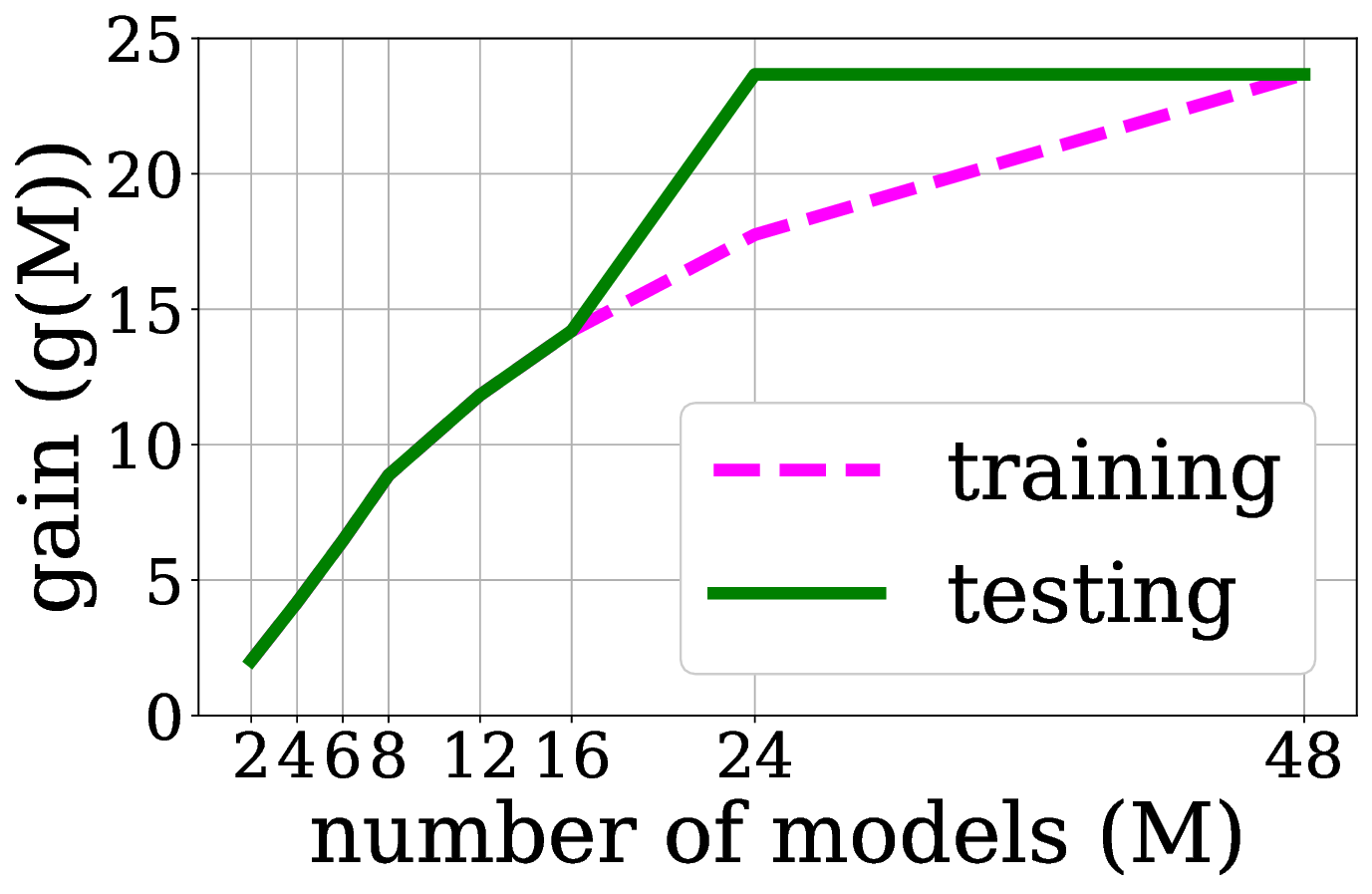}
         \caption{Synthetic(1,1)-A}
         \label{fig:5a}
     \end{subfigure}
     \hfill
     \begin{subfigure}[b]{0.3\textwidth}
         \centering
         \includegraphics[width=\textwidth]{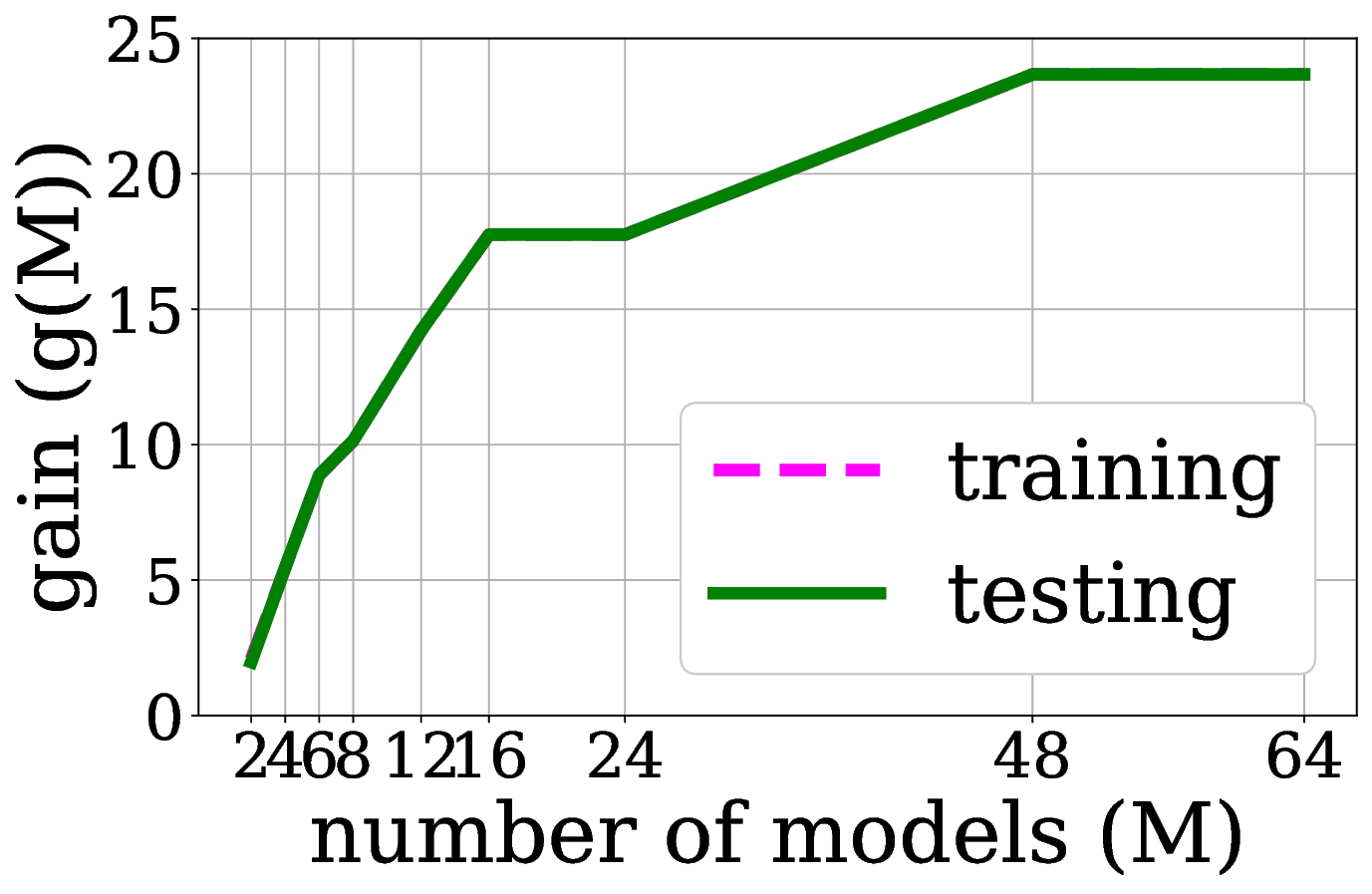}
         \caption{Synthetic(1,1)-B}
         \label{fig:5b}
     \end{subfigure}
     \hfill
     \begin{subfigure}[b]{0.3\textwidth}
         \centering
         \includegraphics[width=\textwidth]{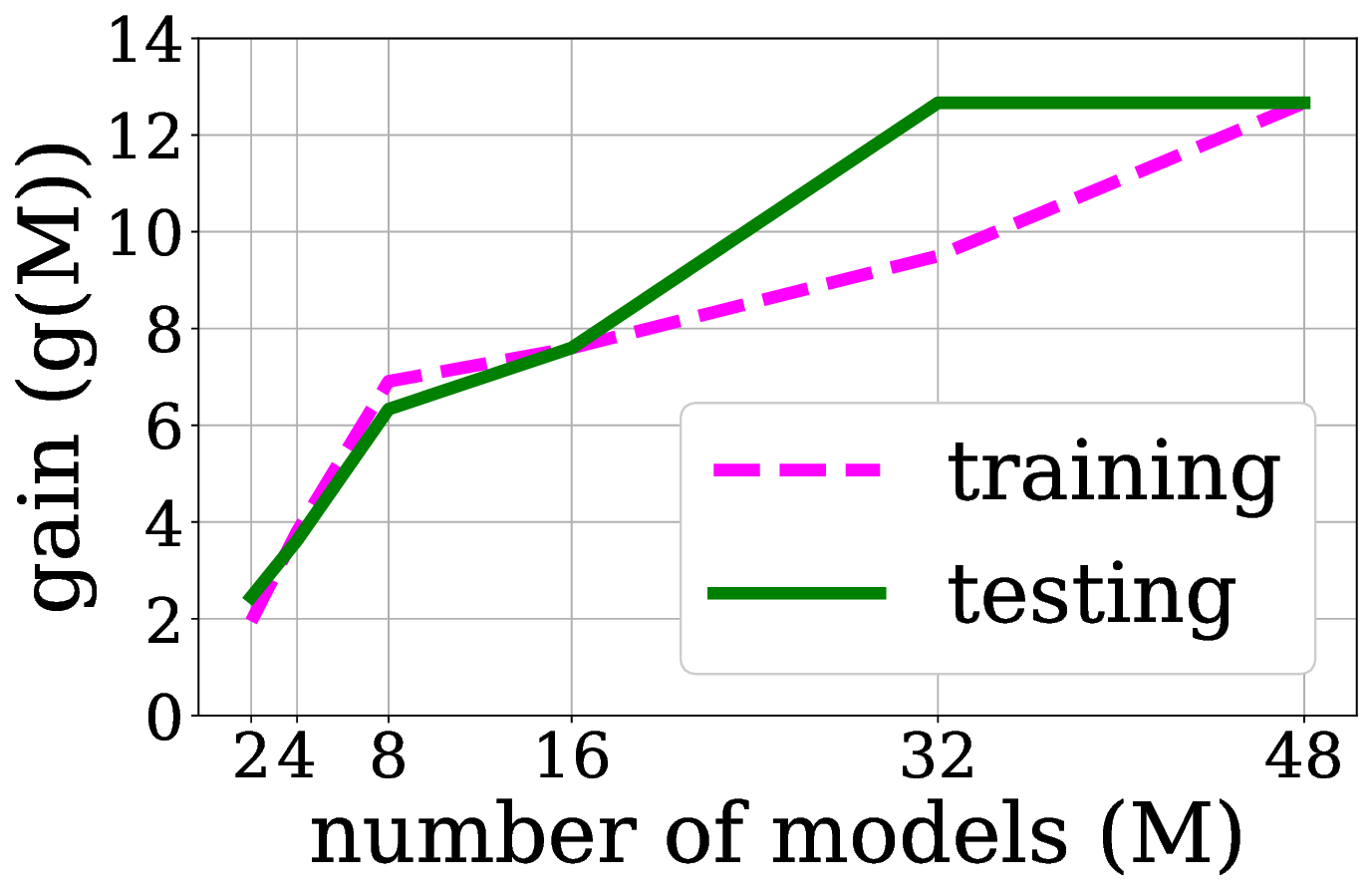}
         \caption{CelebA}
         \label{fig:5c}
     \end{subfigure}
        \caption{Gain vs $M$ for full device participation}
        \label{fig:fig5}
\end{figure}
For full device participation, we observe from Fig. \ref{fig:fig5} that gain increases with $M$ for both Synthetic(1,1) and CelebA datasets with the trend being sub-linear in nature.
\subsubsection{Partial device participation:}
For Synthetic(1,1)-A, we have $N=200$ clients (out of which 32 are sampled every round) and $T_1=200$. During single model FedAvg, we get 61.1\% training accuracy and 61.3\% test accuracy at the end of 200 rounds.

For Synthetic(1,1)-B, we have $N=200$ clients (out of which 32 are sampled every round) and $T_1=200$. During single model FedAvg, we get 58.4\% training accuracy and 57.7\% test accuracy at the end of 200 rounds.

For CelebA, we have 96 clients (out of which 24 are sampled every round) and $T_1 = 75$. We get 78\% training accuracy and 71.5\% test accuracy at the end of 75 rounds of single model FedAvg.

\begin{figure}
     \centering
     \begin{subfigure}[b]{0.3\textwidth}
         \centering
         \includegraphics[width=\textwidth]{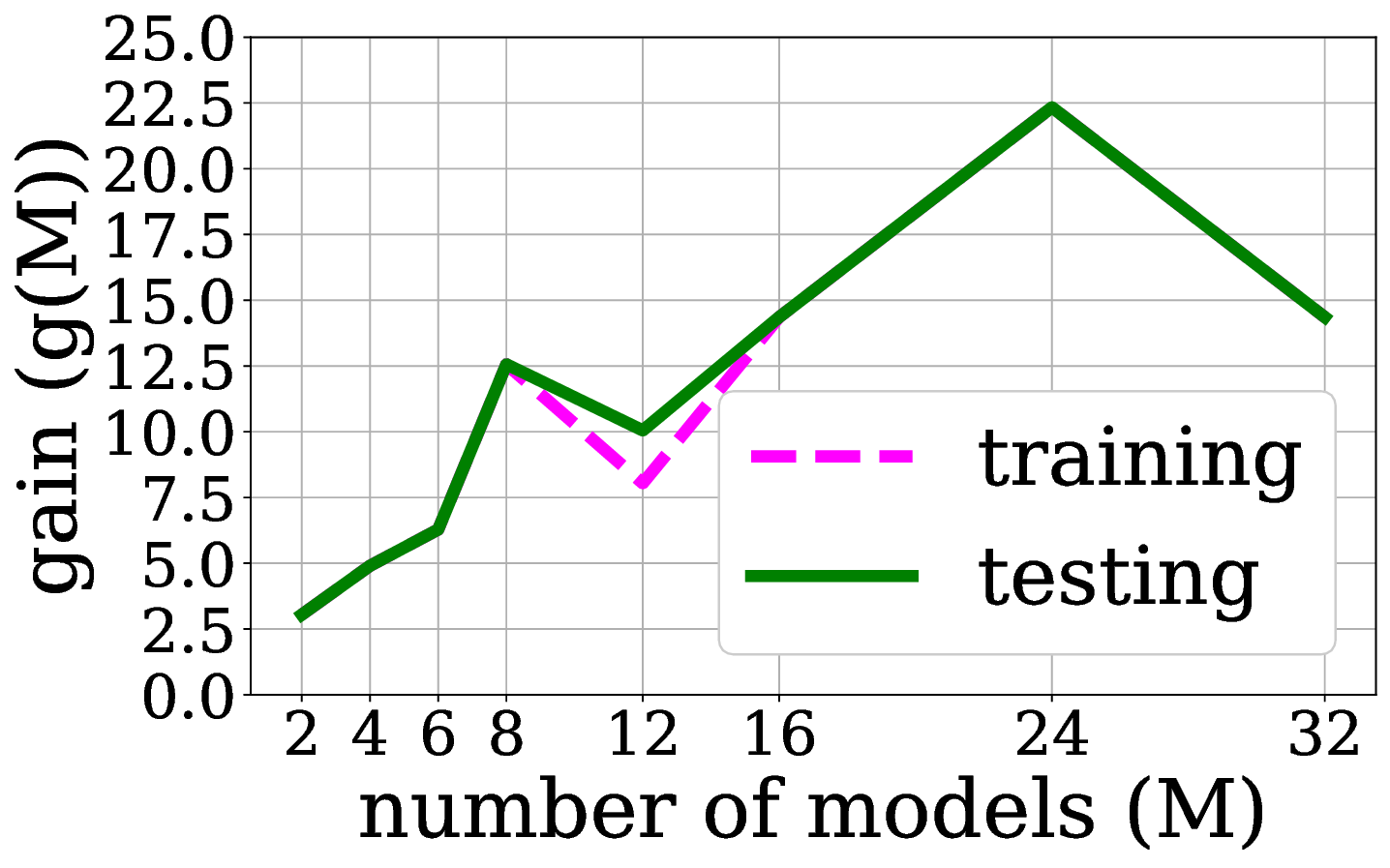}
         \caption{Synthetic(1,1)-A}
         \label{fig:6a}
     \end{subfigure}
     \hfill
     \begin{subfigure}[b]{0.3\textwidth}
         \centering
         \includegraphics[width=\textwidth]{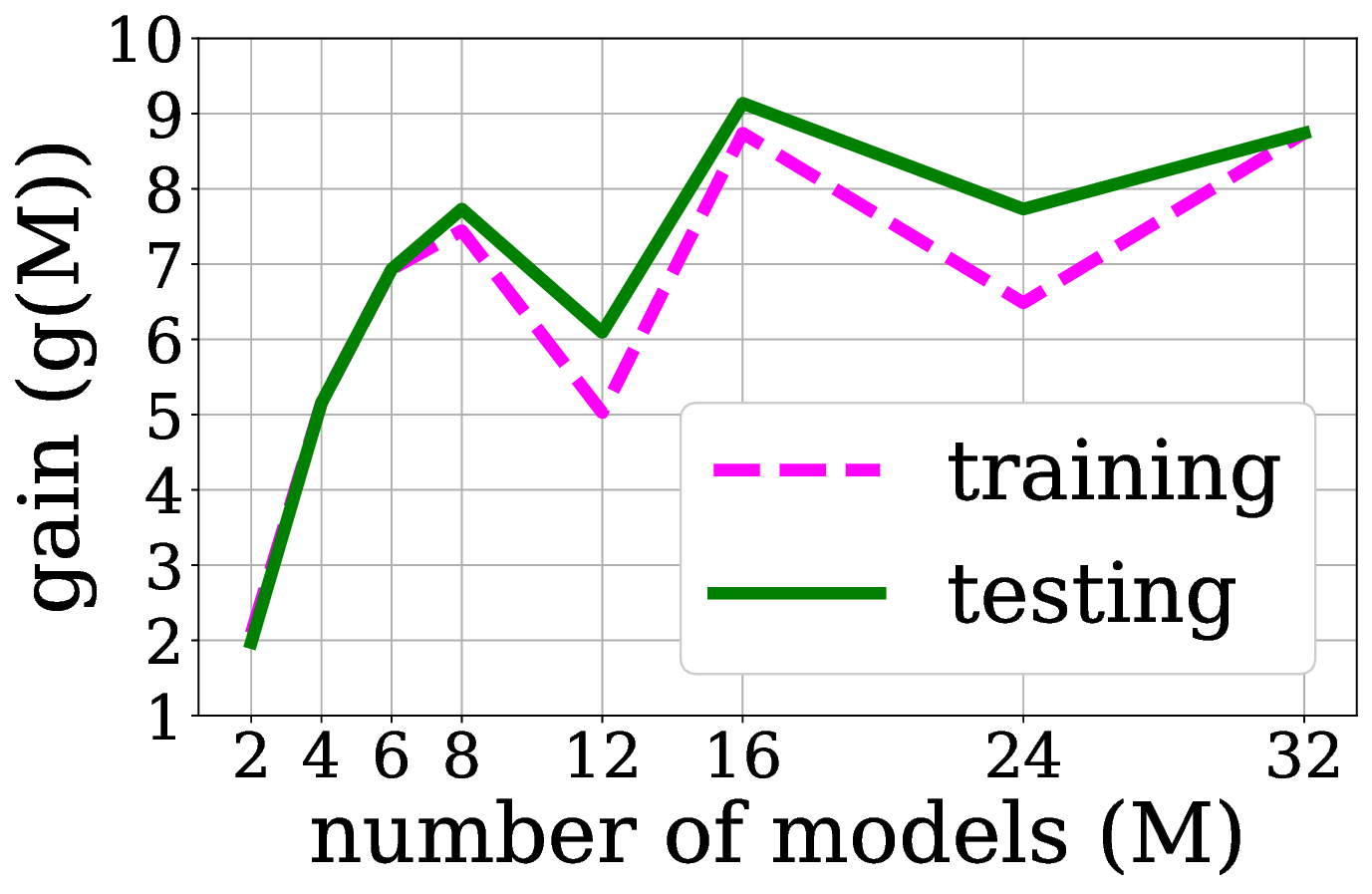}
         \caption{Synthetic(1,1)-B}
         \label{fig:6b}
     \end{subfigure}
     \hfill
     \begin{subfigure}[b]{0.3\textwidth}
         \centering
         \includegraphics[width=\textwidth]{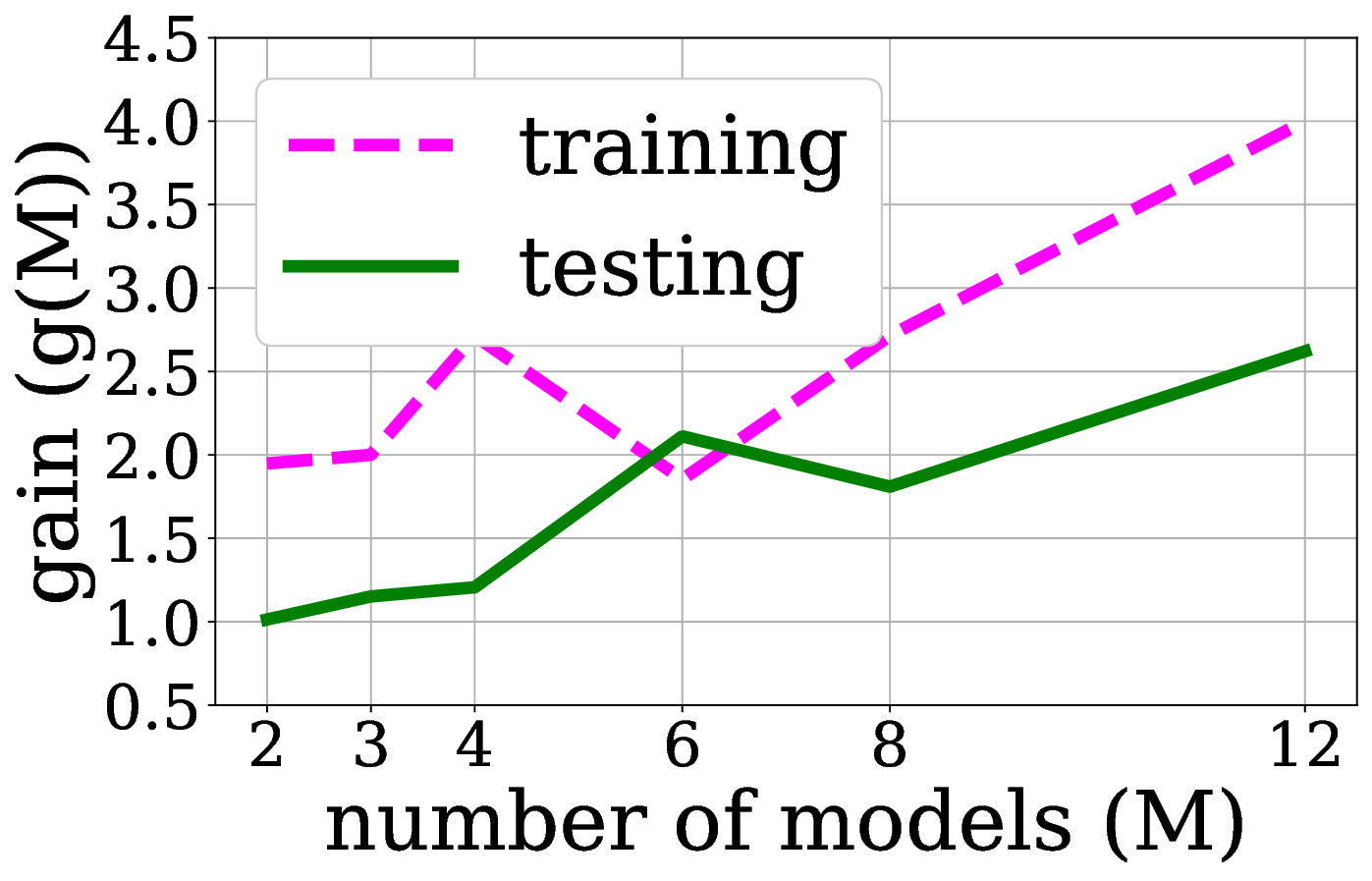}
         \caption{CelebA}
         \label{fig:6c}
     \end{subfigure}
        \caption{Gain vs $M$ for partial device participation}
        \label{fig:fig6}
\end{figure}
When there is partial device participation, for both datasets, we observe in Fig. \ref{fig:fig6} that gain increases with $M$ for the most part while decreasing at some instances. Although, there are some dips, gain is always found to be more than 1.
\begin{remark}
It is important to note that the learning task in CelebA dataset invloves CNNs, rendering it a non-convex nature. This, however, does not severely impact the gain, as we still observe it to always increase with $M$ for full device participation.
\end{remark}
\begin{remark}
Although, Theorem \ref{thm3} assumes full device participation, we see the benefit of multi-model FL in partial device participation scenario. For all three datasets, gain is always found to be significantly greater than 1.
\end{remark}
\subsection{Real-world Scenarios}
We perform two types of real world examples, one involving models that are similar in some aspect and the other involving completely different models. In these experiments, $T_1$ denotes the number of rounds for which single model FedAvg was been run for each model. Further, $T_M$ denotes the number of rounds of multi-model FL, after which each model an accuracy that is at least what was achieved with $T_1$ rounds of single model FedAvg.
\subsubsection{Similar models:} First one tests Theorem \ref{thm3} where each of the $M(=9)$ models is a binary image classification based on a unique facial attribute, using CelebA dataset. Table \ref{tab1} shows the results of our experiment.

Based on the the values of $T_1$ and $T_M$ from Table \ref{tab1}, we have the following for training and testing cases.
\begin{itemize}
    \item Gain in training = $9\times 50/117 = \textbf{3.846}$
    \item Gain in testing = $9\times 50/126 = \textbf{3.571}$
\end{itemize}
\begin{table}
\caption{MFA-RR training 9 different CNN models}
\centering
\begin{tabular}{|c|c|c|c|c|}
\hline
\textbf{Facial attribute} & \multicolumn{2}{|c|}{\textbf{Train Accuracy}}& \multicolumn{2}{|c|}{\textbf{Test Accuracy}} \\
\cline{2-5} 
\textbf{for classification} & $T_1=50$ & $T_M = 117$ & $T_1=50$ & $T_M=126$ \\
\hline
Eyeglasses & 74.7 & 84.6 & 69.1 & 71.9\\
Baldness & 74.0 & 74.5 & 66.8 & 69.4\\
Goatee & 74.3 & 83.5 & 64.7 & 68.7\\
Wearing Necklace & 73.3 & 80.3 & 66.2 & 72.6\\
Smiling & 72.0 & 78.7 & 76.0 & 79.7\\
Moustache & 74.1 & 82.1 & 65.4 & 72.1\\
Male & 77.1 & 85.0 & 58.7 & 63.5\\
Wearing Lipstick & 75.4 & 83.8 & 65.9 & 72.7\\
Double Chin & 75.3 & 83.9 & 63.9 & 69.3\\
\hline
\end{tabular}
\label{tab1}

\end{table}
\subsubsection{Completely different models:} In the second one, we do a mixed model experiment where one model is logistic regression (Synthetic(1,1) with 60 dimensional vectors into 5 classes) and the other model is CNN (binary classification of face images based on presence of eyeglasses).\\
Based on $T_1$ and $T_M$ from Table \ref{tab2}, we get the following values of gain for training and testing cases,
\begin{itemize}
    \item Gain in training = $2 \times 100/100 = \textbf{2.0}$
    \item Gain in testing = $2 \times 100/142 = \textbf{1.41}$
\end{itemize}
\begin{table}
\caption{MFA-RR training logistic regression and convolutional NN simultaneously}
\centering
\begin{tabular}{|c|c|c|c|c|}
\hline
\textbf{Model} & \multicolumn{2}{|c|}{\textbf{Train Accuracy}}& \multicolumn{2}{|c|}{\textbf{Test Accuracy}} \\
\cline{2-5} 
\textbf{Type} & $T_1=100$ & $T_M = 100$ & $T_1=100$ & $T_M=142$ \\
\hline
Logistic Regression & 51.9 & 52.4 & 52.8 & 55.6\\
Convolutional NN & 86.7 & 87.2 & 75.8 & 77.5\\
\hline
\end{tabular}
\label{tab2}
\end{table}
\section{Conclusions}

In this work, we focus on the problem of using Federated Learning to train multiple independent models simultaneously using a shared pool of clients. We propose two variants of the widely studied FedAvg algorithm, in the multi-model setting, called MFA-Rand and MFA-RR, and show their convergence. In case of MFA-RR, we show that an increasing data sample size (for client side SGD iterations), helps improve the speed of convergence greatly $\left(\mathcal{O}\left(\frac{1}{T}\right) \text{ instead of } \mathcal{O}\left(\frac{1}{T^{1/4}}\right)\right)$.

Further, we propose a performance metric to access the advantage of multi-model FL over single model FL. We characterize conditions under which running MFA-Rand for $M$ models simultaneously is advantageous over running single model FedAvg for each model sequentially. We perform experiments in strongly convex and convex settings to corroborate our analytical results. By running experiments in a non-convex setting, we see the benefits of multi-model FL in deep learning. We also run experiments that are out of the scope of the proposed setting. These were the partial device participation experiments and the real world scenarios. Here also we see an advantage in training multiple models simultaneously.

Further extensions to this work include theoretical analysis of partial device participation scenarios, and convergence guarantees, if any, for unbiased client selection algorithms \cite{bhuyan2022multi} in multi-model FL.
\bibliographystyle{splncs04}
\bibliography{refs}

\newpage
\appendix
\section{Appendix}
\subsection{Method of Analysis}
We analyze the multi-model algorithms from the perspective of one of the $M$ models. Proving convergence for one of the models is enough as Assumptions \ref{assump1}, \ref{assump2}, \ref{assump3} and \ref{assump4} hold for all the $M$ models. To that end, below is MFA-Rand and MFA-RR from the perspective of one of the $M$ models.

In addition to that, we are dropping the time index (or frame index) of the set of subsets during analysis of MFA-Rand (or MFA-RR). This is because we are analysing MFA-Rand over one round and MFA-RR over one frame, during which the set of subsets remain fixed.

\subsubsection{MFA-Rand:}
From the perspective of one of the models, it is equivalent to sampling $\frac{N}{M}$ clients out of $N$ clients, every round. We can, therefore, refer to the analysis of single-model FedAvg with partial device participation in \cite{li2019convergence}.

\subsubsection{MFA-RR:}
A model goes over each of the $M$ client-subsets (created at the start of $t = (l-1)M+1$) exactly once during $t = (l-1)M+1$ to $t = (l-1)M+M$. This means that a model goes over each one of $\{\mathcal{S}_1, \mathcal{S}_2,.. \mathcal{S}_M \}$ exactly once during that frame.

\subsection{Convergence of MFA-Rand}
In \cite{li2019convergence}, the convergence of FedAvg for partial device participation is stated as,
\begin{equation}\label{eqn11}
    \mathbb{E}||\overline{\textbf{w}}_t - \textbf{w}_*||^2 \le \frac{\nu}{t+\gamma}
\end{equation}
where,
\begin{equation}\label{eqn12}
    \nu = \max\left\{\frac{\beta^2 (B+C)}{\beta\mu-1},\mathbb{E}||\overline{\textbf{w}}_1 - \textbf{w}_*||^2(1+\gamma)\right\},
\end{equation}
where,
\begin{equation}\label{eqn13}
    B = 6L\Gamma  + \frac{1}{N^2}\sum_{k=1}^N \sigma_k^2 + 8(E-1)^2 G^2.
\end{equation}
where $\sigma_k^2$ is the upper bound on the variance of $\nabla F_k(x,\xi)$ for any size of $\xi$. We can the get its value by setting $|\xi| = 0$ in Lemma 1.\\
Here, $K$ is the number of clients selected per round. Since, we have $K=\frac{N}{M}$, we put that in the expression of $C$ in \cite{li2019convergence} giving us
\begin{equation}\label{eqn14}
    C = \frac{N-K}{(N-1)K}E^2 G^2 = \frac{M-1}{N-1}E^2 G^2.
\end{equation}
One important thing to note is that $\eta_t$ is constant during the $E$ iterations in round $t$. However, \cite{li2019convergence} has a decreasing step size even during the $E$ iterations. This is the reason why one will find a factor of 4 absent in inequality \ref{eqn14} when compared to its counterpart in \cite{li2019convergence}.\\
Using Cauchy-Schwartz theorem, we have,
\begin{equation}\label{eqn15}
    \left(\mathbb{E}||\overline{\textbf{w}}_t - \textbf{w}_*||\right)^2 \le \mathbb{E}||\overline{\textbf{w}}_t - \textbf{w}_*||^2.
\end{equation}
We therefore have,
\begin{equation}\label{eqn16}
    \mathbb{E}||\overline{\textbf{w}}_t - \textbf{w}_*|| \le \frac{\sqrt{\nu}}{\sqrt{t+\gamma}}.
\end{equation}

\subsection{Convergence of MFA-RR}
We start with introducing some new notation only for the purposes of the proof. We first drop the $m$ subscript that indicated model number $m$ as we need to prove convergence for only one of the $M$ models. Below are the notations used frequently in the proof. Some of them are adopted from \cite{igd_ref}.

\begin{enumerate}
    \item $l$: frame number
    \item $i$: $i^{\text{th}}$ round in current frame $(1 \le i \le M)$
    \item Local iteration: stochastic gradient descent iteration at local device
    \item Global iteration: stochastic centralized gradient descent iteration (virtual)
    \item $\mathcal{S}_i$: $i^{\text{th}}$ subset of clients to be used in the $i^{\text{th}}$ round of a frame (this may differ across frames but we analyse MFA-RR over a single frame and hence, do not index it by frame number) 
    \item $\overline{\textbf{w}}_i^l$: global weight $\overline{\textbf{w}}_{\text{MFA-RR},t}$ (subscript $m$ dropped) at $t = (l-1)M + i$
    \item $\textbf{u}_p$: global weight vector (virtual) of $p^{\text{th}}$ centralized full GD iteration from $\overline{\textbf{w}}_1^l$
    \item $\textbf{u}_{k,p}$: local weight vector of $p^{\text{th}}$ local SGD iteration of $k^{\text{th}}$ client. Since a client is used by a model exactly once in a frame, there is no need to index this variable by the round number.
    \item $\alpha_l$: learning rate for all $M$ rounds in $l^{\text{th}}$ frame.
\end{enumerate}
The local update rule is
\begin{equation}\label{eqn17}
    \textbf{u}_{k,p+1} = \textbf{u}_{k,p} - \alpha_l \nabla F_k(\textbf{u}_{k,p}, \xi_{k,p}) \text{  , } \forall \text{ } k \in \mathcal{S}_i \text{ and } p=1,2,..,E
\end{equation}
where $\textbf{u}_{k,1} = \overline{\textbf{w}}_i^l$. Therefore, the update at the $k^{th}$ client is
\begin{equation}
    \textbf{u}_{k,E+1} - \overline{\textbf{w}}_i^l = \alpha_l \sum_{p=1}^E \nabla F_k(\textbf{u}_{k,p}, \xi_{k,p})
\end{equation}
The global update rule involves summing weight updates from clients in $\mathcal{S}_i$ and multiplying a factor of $\frac{1}{N}$ to it.
\begin{equation}\label{eqn18}
    \overline{\textbf{w}}_{i+1}^l = \overline{\textbf{w}}_{i}^l - \frac{1}{N} \sum_{k \in \mathcal{S}_i} \left(\alpha_l \sum_{p=1}^E \nabla F_k(\textbf{u}_{k,p}, \xi_{k,p}) \right)
\end{equation}
Over one frame, the above expression evaluates to
\begin{equation}\label{eqn19}
    \overline{\textbf{w}}_{1}^{l+1} = \overline{\textbf{w}}_{M+1}^{l} = \overline{\textbf{w}}_{1}^{l} - \frac{\alpha_l}{N} \sum_{i=1}^M \sum_{k \in \mathcal{S}_i} \bigg(\sum_{p=1}^E \nabla F_k(\textbf{u}_{k,p}, \xi_{k,p}) \bigg).
\end{equation}
Now we will compare this with $E$ global iterations of centralized GD,
\begin{align*}\label{eqn20}
    \overline{\textbf{w}}_{1}^{l+1} &= \overline{\textbf{w}}_{1}^{l} - \alpha_l \sum_{p=1}^E \nabla F(\textbf{u}_p) \\
    &- \alpha_l  \left( \frac{1}{N} \sum_{i=1}^M \sum_{k \in \mathcal{S}_i} \sum_{p=1}^E \nabla F_k(\textbf{u}_{k,p}) - \sum_{p=1}^E \nabla F(\textbf{u}_p) \right)\\
    &- \frac{\alpha_l}{N} \sum_{i=1}^M \sum_{k \in \mathcal{S}_i} \sum_{p=1}^E \left(\nabla F_k(\textbf{u}_{k,p}, \xi_{k,p}) - \nabla F_k(\textbf{u}_{k,p}) \right) \numberthis
\end{align*}
where $\textbf{u}_1 = \overline{\textbf{w}}_{1}^{l}$ and,
\begin{equation}\label{eqn21}
    \textbf{u}_{p+1} = \textbf{u}_{p} - \alpha_l \nabla F(\textbf{u}_{p}) \text{  , } p=1,2,..,E.
\end{equation}
So,
\begin{equation}\label{eqn22}
    \textbf{u}_{E+1} = \overline{\textbf{w}}_{1}^{l} - \alpha_l \sum_{p=1}^E \nabla F(\textbf{u}_p)
\end{equation}
We define error $\textbf{e}^l$ as,
\begin{align}
    \textbf{e}^l &= \sum_{p=1}^E \nabla F(\textbf{u}_p) - \frac{1}{N} \sum_{i=1}^M \sum_{k \in \mathcal{S}_i} \bigg(\sum_{p=1}^E \nabla F_k(\textbf{u}_{k,p}) \bigg) \label{eqn23}\\
    &= \sum_{p=1}^E \frac{1}{N} \sum_{i=1}^M \sum_{k \in \mathcal{S}_i} \nabla F_k(\textbf{u}_p) - \frac{1}{N} \sum_{i=1}^M \sum_{k \in \mathcal{S}_i} \sum_{p=1}^E \nabla F_k(\textbf{u}_{k,p}) \label{eqn24}\\
    &= \frac{1}{N} \sum_{i=1}^M \sum_{k \in \mathcal{S}_i} \sum_{p=1}^E (\nabla F_k(\textbf{u}_p) - \nabla F_k(\textbf{u}_{k,p}). \label{eqn25}
\end{align}
And we define $\textbf{d}^l$ as
\begin{equation}\label{eqn26}
    d^l = \frac{1}{N} \sum_{i=1}^M \sum_{k \in \mathcal{S}_i} \sum_{p=1}^E \left(\nabla F_k(\textbf{u}_{k,p}, \xi_{k,p}) - \nabla F_k(\textbf{u}_{k,p}) \right).
\end{equation}
Therefore,
\begin{equation}\label{eqn27}
    \overline{\textbf{w}}_{1}^{l+1} = \textbf{u}_{E+1} + \alpha_l \textbf{e}^l - \alpha_l \textbf{d}^l
\end{equation}
We now track the expected distance between $\overline{\textbf{w}}_{1}^{l}$ and $\textbf{w}_*$. Subtracting $\textbf{w}_*$ on both sides of the above equation and taking expectation of norm, we get
\begin{equation}\label{eqn28}
    \mathbb{E}||\overline{\textbf{w}}_{1}^{l+1} -\textbf{w}_*|| \le \mathbb{E}||\textbf{u}_{E+1} - \textbf{w}_*|| + \alpha_l\mathbb{E}||\textbf{e}^l|| +
    \alpha_l \mathbb{E}||\textbf{d}^l||.
\end{equation}
We state lemmas \ref{lemma1}, \ref{lemma2}, \ref{lemma3} and \ref{lemma4} and use them to prove Theorem \ref{thm2}. Proofs of lemmas can be found after proof of Theorem \ref{thm2}.

\begin{lemma}\label{lemma1}
Under assumptions \ref{assump3} and \ref{assump4}, for each client $k$ and sample $\xi_{k,p}$, we have
\begin{equation*}
    var(\nabla F_k(u_{k,p})) \le 4\left(\frac{\mathcal{N} - |\xi_{k,p}|}{\mathcal{N}} \right)^2 (\beta_1 + \beta_2 G^2).
\end{equation*}
\end{lemma}

\begin{lemma}\label{lemma2}
Under assumptions \ref{assump1} and \ref{assump2}, for $E$ iterations of centralized GD with learning rate $\alpha_l \le \frac{1}{L}$, we get
\begin{equation*}
    \left|\left| \textbf{u}_{E+1} - w_* \right|\right| \le (1-\alpha_l\mu)^E||\overline{\textbf{w}}_1^l - \textbf{w}_*||.
\end{equation*}
\end{lemma}

\begin{lemma}\label{lemma3}
Under assumptions \ref{assump2} and \ref{assump3} we get the following bound on $\mathbb{E}||\textbf{e}^l||$ for all $E\ge 1$ and $M \ge 1$
\begin{equation*}
    \mathbb{E}||\textbf{e}^l|| \le \alpha_l LG \bigg(\frac{E^2 (M-1)}{2M} + E(E-1) \bigg).
\end{equation*}
\end{lemma}

\begin{lemma}\label{lemma4}
Under assumptions \ref{assump3} and \ref{assump4}, we get the following bound on $\mathbb{E}||\textbf{d}^l||$ for all $E\ge 1$ and $M \ge 1$
\begin{equation*}
    \mathbb{E}||\textbf{d}^l||  \le 2E \left(\frac{\mathcal{N} - \mathcal{N}_s(l)}{\mathcal{N}} \right) \sqrt{\beta_1 + \beta_2 G^2}
\end{equation*}
where $\mathcal{N}_s(l)$ is the common sample size for local SGD iterations at all clients in frame $l$.
\end{lemma}


\subsubsection{Proof of Theorem 2:} Combining lemmas \ref{lemma1}, \ref{lemma2}, \ref{lemma3} and \ref{lemma4}, we have
\begin{align*}\label{eqn29}
    \mathbb{E}||\overline{\textbf{w}}_{1}^{l+1} -\textbf{w}_*|| \le &  (1-\alpha_l \mu)^E \mathbb{E}||\overline{\textbf{w}}_{1}^{l} -\textbf{w}_*||\\
    &+ \alpha_l^2 LG \left(\frac{E^2 (M-1)}{2M} + E(E-1) \right)\\
    &+ \alpha_l 2E \left(\frac{\mathcal{N} - \mathcal{N}_s(l)}{\mathcal{N}} \right) \sqrt{\beta_1 + \beta_2 G^2}. \numberthis
\end{align*}
Now if we have the following sample size evolution
\begin{equation}\label{eqn30}
    \frac{\mathcal{N} - \mathcal{N}_s(l)}{\mathcal{N}} \le \alpha_l \left(\frac{V}{2E  \sqrt{\beta_1 + \beta_2 G^2}} \right)
\end{equation}
for some $V \ge 0$, we get $\mathbb{E}||\textbf{d}^l|| \le \alpha_l V$. Further $(1-\alpha_l \mu)^E \le 1-\alpha_l \mu$ as $E\ge 1$. We therefore have
\begin{equation}\label{eqn31}
    \mathbb{E}||\overline{\textbf{w}}_{1}^{l+1} -\textbf{w}_*|| \le  (1-\alpha_l \mu) \mathbb{E}||\overline{\textbf{w}}_{1}^{l} -\textbf{w}_*|| + \alpha_l^2 (Y+Z+V)
\end{equation}
where $Y = \frac{LG E^2 (M-1)}{2M}$ and $Z = LG E(E-1)$.\\
We now take $\alpha_l = \frac{\beta}{l + \gamma}$ and show through induction that $\mathbb{E}||\overline{\textbf{w}}_1^l - \textbf{w}_*|| \le \frac{\nu}{l + \gamma}$ for some $\nu > 0$ and some $\gamma \ge 0$.\\
For $l=1$ we want,
\begin{equation}\label{eqn32}
     \frac{\nu}{1 + \gamma} \ge \mathbb{E}||\overline{\textbf{w}}_1^1 - \textbf{w}_*||
\end{equation}
or,
\begin{equation}\label{eqn33}
    \nu \ge (1+\gamma)\mathbb{E}||\overline{\textbf{w}}_1^1 - \textbf{w}_*||.
\end{equation}
Next we show that $\mathbb{E}||\overline{\textbf{w}}_1^l - \textbf{w}_*|| \le \frac{\nu}{l + \gamma}$ implies $\mathbb{E}||\overline{\textbf{w}}_1^{l+1} - \textbf{w}_*|| \le \frac{\nu}{l + 1 + \gamma}$ for a certain $\nu$
\begin{align}
    \mathbb{E}||\overline{\textbf{w}}_{1}^{l+1} -\textbf{w}_*|| &\le \bigg(1-\frac{\beta \mu}{l+\gamma} \bigg) \frac{\nu}{l+\gamma} + \frac{\beta^2 (Y+Z+V)}{(l+\gamma)^2} \label{eqn34}\\
    &= \frac{(l+\gamma - \beta \mu)\nu + \beta^2 (Y+Z+V)}{(l+\gamma)^2} \label{eqn35}\\
    &= \frac{(l+\gamma -1 + 1 - \beta \mu)\nu + \beta^2 (Y+Z+V)}{(l+\gamma)^2} \label{eqn36}\\
    &= \frac{l+\gamma-1}{(l+\gamma)^2}\nu - \bigg(\frac{(\beta \mu -1)\nu - \beta^2(Y+Z+V)}{(l+\gamma)^2}\bigg) \label{eqn37}\\
    &\le \frac{\nu}{l+\gamma + 1} - \bigg(\frac{(\beta \mu -1)\nu - \beta^2(Y+Z+V)}{(l+\gamma)^2}\bigg) \label{eqn38}\\
    &\le \frac{\nu}{l+\gamma + 1}. \label{eqn39}
\end{align}
Inequality \ref{eqn38} holds as $\frac{x-1}{x^2} \le \frac{1}{x+1}$ for $x>-1$ and $l+\gamma> -1$ obviously. Inequality \ref{eqn39} holds when
\begin{equation}\label{eqn40}
    (\beta \mu -1)\nu \ge \beta^2(Y+Z+V)
\end{equation}
or,
\begin{equation}\label{eqn41}
    \nu \ge \frac{\beta^2(Y+Z+V)}{\beta \mu -1}
\end{equation}
with $\beta \ge \frac{1}{\mu}$. We therefore have,
\begin{equation}\label{eqn42}
    \nu = \max \bigg\{\frac{\beta^2(Y+Z+V)}{\beta \mu -1}, (1+\gamma)\mathbb{E}||\overline{\textbf{w}}_1^1 - \textbf{w}_*|| \bigg\}.
\end{equation}
Further, we need $\alpha_l \le \frac{1}{L}$ $\forall$ $l\ge 0$. This is satisfied if $\alpha_1 \le \frac{1}{L}$,
\begin{equation}\label{eqn43}
    \frac{\beta}{1 + \gamma} \le \frac{1}{L}
\end{equation}
or,
\begin{equation}\label{eqn44}
    \gamma \ge \beta L - 1.
\end{equation}
Note that $\beta L - 1 \ge 0$ as $\beta \ge \frac{1}{\mu}$.\\
Putting everything together, we get the following for $\alpha_l = \frac{\beta}{l + \gamma}$ with $\beta \ge \frac{1}{\mu}$ and $\gamma \ge \beta L - 1$,
\begin{equation}\label{eqn45}
    \mathbb{E}||\overline{\textbf{w}}_1^l - \textbf{w}_*|| \le \frac{\nu}{l + \gamma} \text{ } \forall \text{ } l\ge 1.
\end{equation}
Therefore for any $\overline{\textbf{w}}_i^l$,
\begin{align}
    \mathbb{E}||\overline{\textbf{w}}_i^l - \textbf{w}_*|| &= \mathbb{E}||(\overline{\textbf{w}}_i^l - \overline{\textbf{w}}_1^l) + \overline{\textbf{w}}_1^l - \textbf{w}_*|| \label{eqn46}\\
    &\le \mathbb{E}||\overline{\textbf{w}}_i^l - \overline{\textbf{w}}_1^l|| + \mathbb{E}||\overline{\textbf{w}}_{1}^l - \textbf{w}_*|| \label{eqn47}\\
    &\le \frac{\alpha_l (i-1)EG}{M} + \frac{\nu}{l + \gamma} \label{eqn48}\\
    &= \frac{\beta E G ((i-1)/M) + \nu}{l+\gamma} \label{eqn49}\\
    &\le \frac{\beta E G ((M-1)/M) + \nu}{l+\gamma}. \label{eqn50}
\end{align}
Therefore,
\begin{equation}\label{eqn51}
    \mathbb{E}||\overline{\textbf{w}}_i^l - \textbf{w}_*|| \le \frac{\beta E G ((M-1)/M) + \nu}{l+\gamma}.
\end{equation}
We can write $l$ as $ 1 + \floor{\frac{t}{M}}$ and $\overline{\textbf{w}}_i^l$ is $\overline{\textbf{w}}_{\text{MFA-RR},t}^{(m)}$. Since, $\frac{t}{M} - 1$ is less than $\floor{\frac{t}{M}}$, we can replace the latter with the former to give Theorem \ref{thm2}. The denominator will still be always non-negative as $\gamma \ge \beta L-1 \ge 0$.\\
The results hold for any set of disjoint client subsets as we do not use any specification information regarding distribution of clients into $\{\mathcal{S}_1,\mathcal{S}_2,...,\mathcal{S}_M\}$. This means that the results hold even if $\{\mathcal{S}_1,\mathcal{S}_2,...,\mathcal{S}_M\}$ is different for every frame.

\subsubsection{Proof of Lemma \ref{lemma1}:} For any sample $\xi_{k,p}$ of data points in any client $k$, we have
\begin{equation}\label{eqn52}
    \nabla F_k(\textbf{u}_{k,p}, \xi_{k,p}) = \frac{1}{|\xi_{k,p}|} \sum_{h \in \xi_{k,p}} \nabla f_{k,h}(\textbf{u}_{k,p}).
\end{equation}
We therefore have
\begin{align}
    \textbf{v}_{k,p} &= \nabla F_k(\textbf{u}_{k,p}, \xi_{k,p}) - \nabla F_k(\textbf{u}_{k,p}) \label{eqn53}\\
    &= \frac{1}{|\xi_{k,p}|} \sum_{h \in \xi_{k,p}} \nabla f_{k,h}(\textbf{u}_{k,p}) - \frac{1}{\mathcal{N}} \sum_{y =1}^\mathcal{N} \nabla f_{k,y}(\textbf{u}_{k,p}) \label{eqn54}\\
    &= \frac{\mathcal{N} - |\xi_{k,p}|}{\mathcal{N} |\xi_{k,p}|} \sum_{h \in \xi_{k,p}} \nabla f_{k,h}(\textbf{u}_{k,p}) - \frac{1}{\mathcal{N}} \sum_{y \not \in \xi_{k,p}} \nabla f_{k,y}(\textbf{u}_{k,p}). \label{eqn55}
\end{align}
Using assumption \ref{assump4}, we get
\begin{align*}
    ||\textbf{v}_{k,p}||^2 &= \left|\left| \frac{\mathcal{N} - |\xi_{k,p}|}{\mathcal{N} |\xi_{k,p}|} \sum_{h \in \xi_{k,p}} \nabla f_{k,h}(\textbf{u}_{k,p}) - \frac{1}{\mathcal{N}} \sum_{y \not \in \xi_{k,p}} \nabla f_{k,y}(\textbf{u}_{k,p}) \right| \right|^2 \numberthis \label{eqn56}\\
    &\le \left (\frac{\mathcal{N} - |\xi_{k,p}|}{\mathcal{N} |\xi_{k,p}|} \sum_{h \in \xi_{k,p}} ||\nabla f_{k,h}(\textbf{u}_{k,p})|| + \frac{1}{\mathcal{N}} \sum_{y \not \in \xi_{k,p}} ||\nabla f_{k,y}(\textbf{u}_{k,p})|| \right)^2 \numberthis \label{eqn57}\\
    &\le \left (\frac{\mathcal{N} - |\xi_{k,p}|}{\mathcal{N} |\xi_{k,p}|} \sum_{h \in \xi_{k,p}} 1^2 + \frac{1}{\mathcal{N}} \sum_{y \not \in \xi_{k,p}} 1^2 \right) \times \\
    &\left (\frac{\mathcal{N} - |\xi_{k,p}|}{\mathcal{N} |\xi_{k,p}|} \sum_{h \in \xi_{k,p}} ||\nabla f_{k,h}(\textbf{u}_{k,p})||^2 + \frac{1}{\mathcal{N}} \sum_{y \not \in \xi_{k,p}} ||\nabla f_{k,y}(\textbf{u}_{k,p})||^2 \right) \numberthis \label{eqn58}\\
    &\le 4\left(\frac{\mathcal{N} - |\xi_{k,p}|}{\mathcal{N}} \right)^2 (\beta_1 + \beta_2||\nabla F_k(\textbf{u}_{k,p})||^2). \numberthis \label{eqn59}
\end{align*}
Using Assumption \ref{assump3},
\begin{align}
    var(\nabla F_k(u_{k,p})) &= \mathbb{E}||\textbf{v}_{k,p}||^2 \label{eqn60}\\
    &\le 4\left(\frac{\mathcal{N} - |\xi_{k,p}|}{\mathcal{N}} \right)^2 (\beta_1 + \beta_2 \mathbb{E}||\nabla F_k(\textbf{u}_{k,p})||^2) \label{eqn61}\\
    &\le 4\left(\frac{\mathcal{N} - |\xi_{k,p}|}{\mathcal{N}} \right)^2 (\beta_1 + \beta_2 G^2). \label{eqn62}
\end{align}

\subsubsection{Proof of Lemma \ref{lemma2}:}
We have the following update rule for $\textbf{u}_p$,
\begin{equation}\label{eqn63}
    \textbf{u}_{p+1} = \textbf{u}_{p} - \alpha_l \nabla F(\textbf{u}_{p}) \text{  , } p=1,2,..,E.
\end{equation}
We therefore have,
\begin{equation}\label{eqn64}
    ||\textbf{u}_{p+1} - \textbf{w}_*|| = ||\textbf{u}_{p} - \alpha_l \nabla F(\textbf{u}_{p}) - \textbf{w}_*|| \text{  , } p=1,2,..,E.
\end{equation}
Now,
\begin{equation}\label{eqn65}
    \nabla F(\textbf{x}_B) - \nabla F(\textbf{x}_A) = \int_{\textbf{x}_A}^{\textbf{x}_B} \nabla^2 F(\textbf{x}) \textbf{dx}
\end{equation}
where $\textbf{dx}$ is an infinitesimal vector. The above expression is an extension of line integral of gradient,
\begin{equation}\label{eqn66}
    g(\textbf{x}_B) - g(\textbf{x}_A) = \int_{\textbf{x}_A}^{\textbf{x}_B} \nabla g(\textbf{x})^T \textbf{dx}.
\end{equation}
Since $\nabla^2 F = [\nabla g_1 \nabla g_2 ... \nabla g_n]^T$ where $\nabla F = [g_1 g_2 ... g_n]^T$, applying equation \ref{eqn66} on each $g_i$ gives equation \ref{eqn65}. Note that the integral is independent of the path taken from $\textbf{x}_A$ to $\textbf{x}_B$, so any $\textbf{dx}$ works as long as $\int_{\textbf{x}_A}^{\textbf{x}_B} \textbf{dx} = \textbf{x}_B - \textbf{x}_A$\\
Now, putting $\textbf{x}_A = \textbf{w}_*$, $\textbf{x}_B = \textbf{u}_p$, $\textbf{x} = \textbf{w}_* + \tau(\textbf{u}_p - \textbf{w}_*)$, we have
\begin{equation}\label{eqn67}
    \nabla F(\textbf{u}_p) - \textbf{0} = \bigg(\int_{0}^{1} \nabla^2 F(\textbf{w}_* + \tau(\textbf{u}_p - \textbf{w}_*)) d\tau \bigg)(\textbf{u}_p - \textbf{w}_*)
\end{equation}
as $\textbf{dx} = d\tau(\textbf{u}_p - \textbf{w}_*)$. Lets define $A_p$ as,
\begin{equation}\label{eqn68}
    A_p = \int_{0}^{1} \nabla^2 F(\textbf{w}_* + \tau(\textbf{u}_p - \textbf{w}_*)) d\tau.
\end{equation}
We therefore have,
\begin{equation}\label{eqn69}
    \nabla F(\textbf{u}_p) = A_p(\textbf{u}_p - \textbf{w}_*).
\end{equation}
Therefore,
\begin{equation}\label{eqn70}
   \textbf{u}_p - \alpha_l \nabla F(\textbf{u}_p) - \textbf{w}_* = (I - \alpha_l A_p)(\textbf{u}_p - \textbf{w}_*).
\end{equation}
Taking norm on both sides,
\begin{align}
    || \textbf{u}_p - \alpha_l \nabla F(\textbf{u}_p) - \textbf{w}_*|| &= ||(I - \alpha_l A_p)(\textbf{u}_p - \textbf{w}_*)|| \label{eqn71}\\
    &\le ||I - \alpha_l A_p||.||\textbf{u}_p - \textbf{w}_*|| \label{eqn72}\\
    &= \sigma_{max}(I - \alpha_l A_p)||\textbf{u}_p - \textbf{w}_*|| \label{eqn73}\\
    &= \left(\max_{j}|\lambda_j(I - \alpha_l A_p)|\right)||\textbf{u}_p - \textbf{w}_*||, \label{eqn74}
\end{align}
where $\sigma_{max}$ is the spectral norm of a matrix. The last equality is true because $A_p$ is a real symmetric matrix, making $I-\alpha_lA_p$ real symmetric. Here, $\lambda_j (\cdot)$ denotes $j^{\text{th}}$ eigenvalue of the matrix.\\
Since $\mu I \preceq \nabla^2 F \preceq L I$ and semi-definiteness holds under addition, the following is true
\begin{equation}\label{eqn75}
    \int_{0}^{1} \mu I d\tau \preceq \int_{0}^{1} \nabla^2 F(\textbf{w}_* + \tau(\textbf{u}_p - \textbf{w}_*)) d\tau \preceq \int_{0}^{1} L I d\tau
\end{equation}
implying,
\begin{equation}\label{eqn76}
    \mu I \preceq A_p \preceq L I
\end{equation}
and,
\begin{equation}\label{eqn77}
    (1-\alpha_l L) I \preceq I - \alpha_l A_p \preceq (1-\alpha_l \mu) I.
\end{equation}
We would like $I - \alpha_l A_p$ to be always positive semi-definite \cite{igd_ref} and therefore impose $\alpha_l \le \frac{1}{L}$.\\
Now, $I - \alpha_l A_p - (1-\alpha_l \mu) I$ has eigenvalues
$\left\{\lambda_j(I - \alpha_l A_p) - (1-\alpha_l \mu)\right\}$. Since, $I - \alpha_l A_p - (1-\alpha_l \mu) I \preceq 0$, all its eigen values are less than or equal to 0 giving us
\begin{equation}\label{eqn78}
    0 \le \lambda_j(I - \alpha_l A_p) \le (1-\alpha_l \mu) \text{ } \forall \text{ } j.
\end{equation}
Therefore,
\begin{align}
    ||\textbf{u}_p - \alpha_l \nabla F(\textbf{u}_p) - \textbf{w}_*|| &\le \left(\max_{j}|\lambda_j(I - \alpha_l A_p)|\right) ||\textbf{u}_p - \textbf{w}_*|| \label{eqn79}\\
    &\le (1- \alpha_l \mu) ||\textbf{u}_p - \textbf{w}_*||. \label{eqn80}
\end{align}
Putting everything together,
\begin{equation}\label{eqn81}
    ||\textbf{u}_{p+1} - \textbf{w}_*|| \le (1- \alpha_l \mu)||\textbf{u}_p - \textbf{w}_*|| \text{  , } p=1,2,.., E
\end{equation}
which gives us
\begin{align}
    ||\textbf{u}_{E+1} - \textbf{w}_*|| &\le (1-\alpha_l\mu)^E||\textbf{u}_{1} - \textbf{w}_*|| \label{eqn82}\\
    &= (1-\alpha_l\mu)^E||\overline{\textbf{w}}_1^l - \textbf{w}_*||. \label{eqn83}
\end{align}
Therefore, 
\begin{equation}\label{eqn84}
    \mathbb{E}||\textbf{u}_{E+1} - \textbf{w}_*|| \le (1-\alpha_l\mu)^E \mathbb{E} ||\overline{\textbf{w}}_1^l - \textbf{w}_*||.
\end{equation}
\subsubsection{Proof of Lemma \ref{lemma3}:}
We have norm of $\textbf{e}^l$ as
\begin{align}
    \mathbb{E}||\textbf{e}^l|| &= \bigg|\bigg|\frac{1}{N} \sum_{i=1}^M \sum_{k \in S_i} \sum_{p=1}^E (\nabla F_k(\textbf{u}_p) - \nabla F_k(\textbf{u}_{k,p}))\bigg|\bigg| \label{eqn85}\\
    &\le \frac{1}{N} \sum_{i=1}^M \sum_{k \in S_i} \sum_{p=1}^E \mathbb{E}||\nabla F_k(\textbf{u}_p) - \nabla F_k(\textbf{u}_{k,p})||. \label{eqn86}
\end{align}
From assumption \ref{assump2}, $F_k$ is $L$-smooth, giving us
\begin{equation}\label{eqn87}
    ||\nabla F_k(\textbf{u}_p) - \nabla F_k(\textbf{u}_{k,p})|| \le L||\textbf{u}_p - \textbf{u}_{k,p}||
\end{equation}
We now find an upper bound on $||\textbf{u}_p - \textbf{u}_{k,p}||$,
\begin{align}
    ||\textbf{u}_p - \textbf{u}_{k,p}|| &= \bigg|\bigg| \sum_{j=1}^{p-1} (\textbf{u}_{j+1}-\textbf{u}_j) + \textbf{u}_1 - \textbf{u}_{k,1} + \sum_{j=1}^{p-1} (\textbf{u}_{k,j}-\textbf{u}_{k,j+1}) \bigg|\bigg| \label{eqn88}\\
    &\le \sum_{j=1}^{p-1} ||\textbf{u}_{j+1}-\textbf{u}_j|| + ||\textbf{u}_1 - \textbf{u}_{k,1}|| + \sum_{j=1}^{p-1} ||\textbf{u}_{k,j}-\textbf{u}_{k,j+1}|| \label{eqn89}\\
    &\le \sum_{j=1}^{p-1} ||\alpha_l \nabla F(\textbf{u}_j)|| + ||\textbf{u}_1 - \textbf{u}_{k,1}|| + \sum_{j=1}^{p-1} ||\alpha_l \nabla F_k(\textbf{u}_{k,j}, \xi_{k,j})||. \label{eqn90}
\end{align}
Therefore,
\begin{align*}
    \mathbb{E}||\textbf{u}_p - \textbf{u}_{k,p}|| \le& \alpha_l \sum_{j=1}^{p-1} \left(\mathbb{E}|| \nabla F(\textbf{u}_j)|| + \mathbb{E}||\nabla F_k(\textbf{u}_{k,j}, \xi_{k,j})|| \right) + \mathbb{E}||\textbf{u}_1 - \textbf{u}_{k,1}|| \numberthis \label{eqn91}\\
    \le& \alpha_l \sum_{j=1}^{p-1} \left(|| \nabla F(\textbf{u}_j)|| + \sqrt{\mathbb{E}||\nabla F_k(\textbf{u}_{k,j}, \xi_{k,j})||^2} \right) \\
    &+ \mathbb{E}||\textbf{u}_1 - \textbf{u}_{k,1}|| \numberthis \label{eqn92}\\
    \le& \alpha_l \sum_{j=1}^{p-1}  (|| \nabla F(\textbf{u}_j)|| + G) + \mathbb{E}||\textbf{u}_1 - \textbf{u}_{k,1}|| \numberthis \label{eqn93}\\
    \le& 2(p-1) \alpha_l G + \mathbb{E}||\textbf{u}_1 - \textbf{u}_{k,1}|| \numberthis \label{eqn94}\\
    =&  2(p-1) \alpha_l G + \mathbb{E}||\overline{\textbf{w}}_{1}^{l} - \overline{\textbf{w}}_i^l||. \numberthis \label{eqn95}
\end{align*}
Inequality \ref{eqn94} holds as $||\nabla F(\textbf{x})|| = ||(1/N)\sum_{k=1}^N \nabla F_k(\textbf{x})|| \le $\\ $(1/N)\sum_{k=1}^N ||\nabla F_k(\textbf{x})||  = (1/N)\sum_{k=1}^N ||\mathbb{E}\nabla F_k(\textbf{x}, \xi)|| \le$\\ $(1/N)\sum_{k=1}^N \mathbb{E}||\nabla F_k(\textbf{x},\xi)|| \le G$\\
Inserting inequality \ref{eqn87} and \ref{eqn95} in inequality \ref{eqn86},
\begin{align}
    \mathbb{E}||\textbf{e}^l|| &\le \frac{1}{N} \sum_{i=1}^M \sum_{k \in S_i} \sum_{p=1}^E L(2(p-1) \alpha_l G + \mathbb{E}||\overline{\textbf{w}}_{1}^{l} - \overline{\textbf{w}}_i^l||) \label{eqn96}\\
    &= \frac{L}{N} \sum_{i=1}^M \sum_{k \in S_i} 2\frac{E(E-1)}{2} \alpha_l G + E \mathbb{E} ||\overline{\textbf{w}}_{1}^{l} - \overline{\textbf{w}}_i^l|| \label{eqn97}\\
    &= \frac{L}{N} \sum_{i=1}^M \frac{N}{M} (E(E-1)\alpha_l G + E \mathbb{E} ||\overline{\textbf{w}}_{1}^{l} - \overline{\textbf{w}}_i^l||) \label{eqn98}\\
    &= \bigg(\frac{LE}{M} \sum_{i=1}^M \mathbb{E}||\overline{\textbf{w}}_{1}^{l} - \overline{\textbf{w}}_i^l|| \bigg) + \alpha_l LG E(E-1). \label{eqn99}
\end{align}

We will now get an upper bound on $\mathbb{E}||\overline{\textbf{w}}_{1}^{l} - \overline{\textbf{w}}_i^l||$,

\begin{align}
    \mathbb{E}||\overline{\textbf{w}}_{1}^{l} - \overline{\textbf{w}}_i^l|| &= \mathbb{E}\bigg|\bigg|\sum_{j=1}^{i-1} (\overline{\textbf{w}}_{j}^{l} - \overline{\textbf{w}}_{j+1}^{l}) \bigg|\bigg| \label{eqn100}\\
    &\le \sum_{j=1}^{i-1} \mathbb{E}||\overline{\textbf{w}}_{j}^{l} - \overline{\textbf{w}}_{j+1}^{l}|| \label{eqn101}\\
    &=  \sum_{j=1}^{i-1} \mathbb{E}\bigg|\bigg| \frac{1}{N} \sum_{k \in S_i} \bigg(\alpha_l \sum_{p=1}^E \nabla F_k(\textbf{u}_{k,p}, \xi_{k,p}) \bigg) \bigg|\bigg| \label{eqn102}\\
    &\le \frac{\alpha_l}{N} \sum_{j=1}^{i-1} \sum_{k \in S_i} \sum_{p=1}^E \mathbb{E}||\nabla F_k(\textbf{u}_{k,p}, \xi_{k,p})|| \label{eqn103}\\
    &\le \frac{\alpha_l}{N} \sum_{j=1}^{i-1} \sum_{k \in S_i} \sum_{p=1}^E \sqrt{\mathbb{E}||\nabla F_k(\textbf{u}_{k,p}, \xi_{k,p})||^2} \label{eqn104}\\
    &\le \frac{\alpha_l}{N} \sum_{j=1}^{i-1} \sum_{k \in S_i} \sum_{p=1}^E G \label{eqn105}\\
    &= \alpha_l \times \frac{1}{N} \times (i-1) \times \frac{N}{M} \times EG \label{eqn106}\\
    &= \frac{\alpha_l (i-1)EG}{M}. \label{eqn107}
\end{align}
Inserting inequality \ref{eqn107} in inequality \ref{eqn99},
\begin{align}
    \mathbb{E}||\textbf{e}^l|| &\le \frac{LE}{M} \sum_{i=1}^M \frac{\alpha_l (i-1)EG}{M} + \alpha_l LG E(E-1) \label{eqn108}\\
    &= \frac{\alpha_l LE^2G}{M^2} \times \frac{M(M-1)}{2} + \alpha_l LG E(E-1) \label{eqn109}\\
    &= \alpha_l LG \left(\frac{E^2 (M-1)}{2M} + E(E-1) \right). \label{eqn110}
\end{align}
We finally have following upper bound on norm of error $\textbf{e}^l$,
\begin{equation}
    \mathbb{E}||\textbf{e}^l|| \le \alpha_l LG \left(\frac{E^2 (M-1)}{2M} + E(E-1) \right). \label{eqn111}
\end{equation}

\subsubsection{Proof of Lemma \ref{lemma4}:}
We have the following for $\mathbb{E}||d^l||$.
\begin{align}
    \mathbb{E}||d^l|| &= \mathbb{E} \left|\left|\frac{1}{N} \sum_{i=1}^M \sum_{k \in \mathcal{S}_i} \sum_{p=1}^E \left(\nabla F_k(\textbf{u}_{k,p}, \xi_{k,p}) - \nabla F_k(\textbf{u}_{k,p}) \right) \right|\right| \label{eqn112}\\
    &\le \frac{1}{N} \sum_{i=1}^M \sum_{k \in \mathcal{S}_i} \sum_{p=1}^E \mathbb{E}||\nabla F_k(\textbf{u}_{k,p}, \xi_{k,p}) - \nabla F_k(\textbf{u}_{k,p})|| \label{eqn113}\\
    &\le \frac{1}{N} \sum_{i=1}^M \sum_{k \in \mathcal{S}_i} \sum_{p=1}^E \sqrt{\mathbb{E}||\nabla F_k(\textbf{u}_{k,p}, \xi_{k,p}) - \nabla F_k(\textbf{u}_{k,p})||^2} \label{eqn114}\\
    & \le \frac{1}{N} \sum_{i=1}^M \sum_{k \in \mathcal{S}_i} \sum_{p=1}^E \sqrt{var(\nabla F_k(\textbf{u}_{k,p}))} \label{eqn115}\\
    &\le \frac{1}{N} \sum_{i=1}^M \sum_{k \in \mathcal{S}_i} \sum_{p=1}^E \frac{2(\mathcal{N} - |\xi_{k,p}|)}{\mathcal{N}} \sqrt{\beta_1 + \beta_2 G^2} \label{eqn116}\\
    &= 2E \left(\frac{\mathcal{N} - \mathcal{N}_s(l)}{\mathcal{N}} \right) \sqrt{\beta_1 + \beta_2 G^2}, \label{eqn117}
\end{align}
where, $\mathcal{N}_s(l)$ is the common sample size for local SGD iterations at all clients in frame $l$.

\subsection{Advantage of MFA-Rand over running FedAvg $M$ times}
We defined gain of MFA-Rand over FedAvg as
\begin{equation}\label{eqn118}
    g_{\text{MFA-Rand}}(M,\epsilon) = \frac{MT_1(\epsilon)}{T_{\text{MFA-Rand}}(M,\epsilon)}.
\end{equation}
Here, we make use of the assumption that all $M$ models are of similar complexity. Therefore, when MFA-Rand is used, all models will reach $\epsilon$-accuracy in roughly the same number of rounds. $T_{\text{MFA-Rand}}(M,\epsilon)$ will be same as the number rounds each model took to reach accuracy of $\epsilon$. We use Theorem \ref{thm1} to calculate $T_{\text{MFA-Rand}}(M,\epsilon)$ and drop the $m$ superscript during analysis as number rounds each model takes to reach accuracy of $\epsilon$ is same. At the end, however, we add the $m$ superscript for completeness of proof.\\
We will show that $g_{\text{MFA-Rand}}(1,\epsilon) = 1$ and $\frac{d}{dM} g_{\text{MFA-Rand}}(1,\epsilon) \ge 0$ $\forall$ $M\ge 1$
\subsubsection{Proof of Theorem 3:} In the proof use $T_1$ for $T_1(\epsilon)$, $T_M$ for $T_{\text{MFA-Rand}}(M,\epsilon)$ and $g$ for $g_{\text{MFA-Rand}}(M,\epsilon)$.
\begin{align}
    \frac{d}{dM}g &= \frac{d}{dM} \left(\frac{MT_1}{T_M} \right) \label{eqn119}\\
    &= \frac{T_1 T_M - MT_1 \frac{dT_M}{dM}}{T_M^2} \label{eqn120}\\
    &= \frac{T_1\left(T_M - M\frac{dT_M}{dM}\right)}{T_M^2}. \label{eqn121}
\end{align}
So for $\frac{d}{dM}g > 0$, we need,
\begin{equation} \label{eqn122}
    \frac{T_M}{M} > \frac{dT_M}{dM}.
\end{equation}
We have following convergence result for MFA-Rand (we use $\Delta_t$ as a shorthand for $\Delta^{(m)}_{\text{MFA-Rand}}(t)$),
\begin{equation} \label{eqn123}
    \Delta_{T_M} \le \frac{\sqrt{\nu}}{\sqrt{T_M+\gamma}},
\end{equation}
where, $\nu = \max \left \{ \frac{\beta^2 (B+C)}{\beta \mu - 1}, \mathbb{E}||\overline{\textbf{w}}_{\text{MFA-Rand},1} - \textbf{w}_*||^2(1+\gamma) \right\}$. Note that we drop the $m$ superscript and use convergence of only one model as rest of the models will behave similarly and reach $\epsilon$ accuracy in the same time.
\begin{equation}
    \Delta_{T_M} \le \frac{1}{\sqrt{T_M+\gamma}}\sqrt{\frac{\beta^2 (B+C)}{\beta \mu - 1} + \mathbb{E}||\overline{\textbf{w}}_{\text{MFA-Rand},1} - \textbf{w}_*||^2(1+\gamma)} \label{eqn124}
\end{equation}
For $\Delta_{T_M} \le \epsilon$, satisfying the following equation is sufficient
\begin{equation} \label{eqn126}
    \sqrt{\frac{\frac{\beta^2 (B+C)}{\beta \mu - 1} + \mathbb{E}||\overline{\textbf{w}}_{\text{MFA-Rand},1} - \textbf{w}_*||^2(1+\gamma)}{T_M+\gamma}} = \epsilon.
\end{equation}
Implies,
\begin{equation} \label{eqn127}
    T_M = \frac{\frac{\beta^2 (B+C)}{\beta \mu - 1} + \mathbb{E}||\overline{\textbf{w}}_{\text{MFA-Rand},1} - \textbf{w}_*||^2(1+\gamma) - \epsilon^2 \gamma}{\epsilon^2}. 
\end{equation}
Putting $T_M$ in inequality \ref{eqn122}, we get
\begin{equation} \label{eqn130}
    \frac{\frac{\beta^2 (B+C)}{\beta \mu - 1} + \mathbb{E}||\overline{\textbf{w}}_{\text{MFA-Rand},1} - \textbf{w}_*||^2(1+\gamma) - \epsilon^2 \gamma}{M\epsilon^2} > \frac{\beta^2}{(\beta \mu - 1)\epsilon^2} \frac{dC}{dM}.
\end{equation}
Satisfying the following inequality is sufficient if $\epsilon^2  < (\mathbb{E}||\overline{\textbf{w}}_{\text{MFA-Rand},1} - \textbf{w}_*||)^2 < \mathbb{E}||\overline{\textbf{w}}_{\text{MFA-Rand},1} - \textbf{w}_*||^2$,
\begin{equation} \label{eqn131}
    \frac{B+C}{M} > \frac{dC}{dM}.
\end{equation}
Putting $C = \frac{M-1}{N-1}E^2 G^2$ in the above inequality, the required condition is
\begin{equation} \label{eqn132}
    B > \frac{E^2 G^2}{N-1}.
\end{equation}
Now putting expression of $B$ in the above equation, the following inequality needs to satisfied for increasing gain,
\begin{equation} \label{eqn133}
    6L\Gamma + (1/N^2)\sum_{k=1}^N \sigma_k^2 + 8(E-1)^2 G^2 > \frac{E^2 G^2}{N-1}.
\end{equation}
Now, if $N \ge 2$ (which is required for $M>1$ as a client can train only one model at once),
\begin{equation} \label{eqn134}
    {N-1} > \frac{E^2}{8(E-1)^2} \text{ } \forall \text{ } E\ge 2,
\end{equation}
as the right hand side of the above inequality has maximum value of $\frac{1}{2}$. Therefore,
\begin{equation} \label{eqn135}
    8(E-1)^2 G^2 > \frac{E^2 G^2}{N-1} \text{ } \forall \text{ } E \ge 2.
\end{equation}
Therefore, we have
\begin{equation} \label{eqn136}
     6L\Gamma + (1/N^2)\sum_{k=1}^N \sigma_k^2 + 8(E-1)^2 G^2 > \frac{E^2 G^2}{N-1} \text{ } \forall \text{ } E \ge 2,
\end{equation}
proving that $\frac{d}{dM} g>0$ $\forall$ $E\ge 2$.\\
For inequality \ref{eqn133} to hold for $E=1$, the following is sufficient
\begin{equation} \label{eqn137}
    6L\Gamma > \frac{E^2 G^2}{N-1},
\end{equation}
or
\begin{equation} \label{eqn138}
    N > 1 + \frac{G^2}{6L\Gamma}.
\end{equation}
Since, $T_M = T_1$ when $M=1$, we get that $g = \frac{1\times T_1}{T_1} = 1$ when $M=1$. For $M>1$, under the conditions of $\frac{d}{dM} g > 0$, we have $g > 1$ and increasing $\forall$ $M > 1$.\\
Inequality \ref{eqn138} needs to be satisfied for all models. Therefore, we need the following as the condition for increasing gain for $E=1$,
\begin{equation}
    N > 1 + \max_{m} \left \{ \frac{6L\Gamma^{(m)}}{(G^{(m)})^2}\right \}.
\end{equation}

%
%
%
%




\end{document}